\documentclass{article}

\PassOptionsToPackage{numbers, compress}{natbib}
\usepackage[preprint]{neurips_2026}

\usepackage[utf8]{inputenc} % allow utf-8 input
\usepackage[T1]{fontenc}    % use 8-bit T1 fonts
\usepackage{hyperref}       % hyperlinks
\usepackage{url}            % simple URL typesetting
\usepackage{booktabs}       % professional-quality tables
\usepackage{amsfonts}       % blackboard math symbols
\usepackage{nicefrac}       % compact symbols for 1/2, etc.
\usepackage{microtype}      % microtypography
\usepackage{xcolor}         % colors
\usepackage{amsmath}
\usepackage{multirow}
\usepackage{graphicx}
\usepackage[table]{xcolor}
\usepackage{fvextra}
\usepackage{float}
\usepackage{subcaption}
\usepackage[rightcaption]{sidecap} % in the preamble

\newif\ifshowcomments

\showcommentsfalse  % Use \showcommentsfalse to hide everything

\newcommand{\lm}[1]{\textsc{#1}}

\ifshowcomments
    \newcommand{\guang}[1]{\textcolor{cyan}{[\textbf{#1} -- Guang]}}
    \newcommand{\brian}[1]{\textcolor{purple}{[\textbf{#1} -- Brian]}}
    \newcommand{\vcomment}[1]{\textcolor{orange}{[\textbf{#1} -- Victoria]}}
    \newcommand{\nascomment}[1]{\textcolor{blue}{[\textbf{#1} -- Noah]}}
\else
    \newcommand{\guang}[1]{}
    \newcommand{\brian}[1]{}
    \newcommand{\vcomment}[1]{}
    \newcommand{\nascomment}[1]{}
\fi

\title{LEGATO 2: Toward Multimodal Sheet Music \\ Recognition and Understanding}

\author{%
  Guang Yang$^{\spadesuit}$\thanks{Equal contribution.}
  \quad
  Brian Siyuan Zheng$^{\spadesuit \ast}$
  \quad
  Victoria Ebert$^\spadesuit$
  \quad
  Noah A. Smith$^{\spadesuit, \clubsuit}$ \\
  $^\spadesuit$Paul G. Allen School of Computer Science \& Engineering, University of Washington \\
  $^\clubsuit$Allen Institute for AI \\
  \texttt{\{gyang1, nasmith\}@cs.washington.edu}
}

\begin{document}

\maketitle
\setcounter{footnote}{0} % Avoid increasing the footnote counter caused by thanks

\begin{abstract}
We propose a novel pipeline, Legato~2, for extracting symbolic notation and semantic knowledge from images of sheet music. Legato~2 features the first large-scale neural model for optical music recognition (OMR) to operate sequentially on a system-by-system basis, following the horizontal lines of notation as they are read on the page, rather than treating the page as an undifferentiated image, enabling better scaling to arbitrarily long inputs. It is also the first OMR model capable of generating symbolic transcriptions that include embedded textual content, such as titles and annotations.
The pipeline combines system-level segmentation with an autoregressive vision-LM to capture both local notation details and score structure. 
Across multiple datasets, Legato~2 consistently outperforms prior state of the art. 
We also show that symbolic transcriptions complement visual inputs for frontier language models, improving their interpretation of dense musical documents.
Legato~2 establishes new state-of-the-art performance in both OMR and downstream sheet music understanding. 
%We will release data and code upon publication.
\end{abstract}

\section{Introduction}

Written music is central to many musical traditions as a medium of creation, transmission, study, and performance. In traditions where sheet music is an authoritative artifact,\footnote{Such traditions include Western classical music, many forms of jazz, liturgical music, theater music, and more.} musicians cultivate a specifically \emph{visual} form of musical intelligence: they read spatial arrangements of staves, symbols, lyrics, dynamics, articulations, rehearsal marks, annotations, and page layout, transforming musical notation into coordinated musical action. Scholars likewise read sheet music as a document of musical thought, tracing form, harmony, rhythm, text-setting, instrumentation, genre, influence, and performance practice.  As \citet{Aucoin2025ClassicalMusic} argues, Western classical music in particular is arguably better understood by its ``writtenness'' rather than by a stable sound or style.  Sheet music is thus both a site of creativity and a shared technology of transmission across otherwise different musical idioms.

We take that visual-musical intelligence as a computational challenge. 
We introduce Legato~2, a pipeline for sheet music recognition and understanding: extracting symbolic notation and semantic knowledge from visual musical documents. We evaluate the system both as a stand-alone optical music recognition (OMR) model and as an upstream context provider, feeding symbolic output to a frontier language model for musical question answering.

%Musical Score Recognition and Understanding has been an important research topic in the field of Music Information Retrieval~(MIR). This task consists of two stages: recognizing musical scores from images, and retrieving knowledge from the recognized content. The former stage is usually referred to as Optical Music Recognition~(OMR), the musical analogue of OCR. Previous research in score recognition has primarily focused on OMR, as it is the bedrock of understanding musical scores \vcomment{more of a question than a suggestion: is score recognition and understanding always done with images, never already digitized scores?} \guang{some benchmarks also prompt models with ABC notation for question answering. should we rename it as ``visual musical score understanding''?}. However, with recent advances in large language models~(LLMs), researchers have started to build large models to directly understand musical scores and propose benchmarks to evaluate the capabilities of proprietary models in this domain.

%Previous methods for OMR follow two different paradigms: the classical paradigm splits the task into low-level subtasks and utilizes rule-based methods or lightweight models to solve each~\citep{pipeline_omr_2001, pipeline_omr_2012, omr_review}, whereas the end-to-end paradigm builds large neural models to directly generate the symbolic score~\citep{legato1, smtpp, smt, olimpic}. 
Conventional approaches to OMR typically decompose the task into sequential stages, addressing each with independent modules~\citep{pipeline_omr_2001, pipeline_omr_2012, omr_review}. Because these pipelined architectures inherently suffer from compounding error propagation, recent research has shifted toward integrated solutions. This includes bridging intermediate processing stages~\citep{completeomr} or developing end-to-end models either from-scratch~\citep{smt, smtpp, olimpic} or by finetuning~\citep{legato1}.
The capabilities of large models have made the end-to-end paradigm highly robust, especially for real printed sheet music. 
However, previous methods either focus on a single musical system~\citep{olimpic, smt} or process a full page at once, ignoring music's sequential structure~\citep{smtpp, legato1}. They also do not recognize embedded text such as titles, authorship, and annotations. We therefore build a large neural OMR model that reads system by system, recognizes embedded text, and scales naturally to long documents (see \S\ref{subsec:multi-page}).

The resulting pipeline combines modular and end-to-end components. A vision model segments sheet music into systems, an autoregressive recognition model transcribes each system end-to-end, and a rule-based ABC conversion step merges the outputs. Figure \ref{fig:overview} gives the full inference pipeline.

Beyond OMR, sheet music \emph{understanding} has emerged as a challenging task for vision-language models, which are mainly trained on natural images.  Prior work shows that frontier VLMs fail on most sheet-music questions~\citep{musixqa, ssmr, wildscore}, especially those involving musical semantics.  Rather than modifying these models, we use OMR to provide external symbolic music context.
We find that this approach significantly advances the state of the art in sheet music understanding (\S\ref{subsec:smu}).

Our main contributions are:
\begin{itemize}
    \item We introduce Legato~2, an OMR model for system-by-system recognition (\S\ref{sec:methods}), achieving state-of-the-art performance across test datasets and robust long-document processing (\S\ref{subsec:page-level}, \S\ref{subsec:multi-page}).
    \item Legato~2 is the first neural OMR architecture to transcribe embedded text alongside musical notation, capturing titles, composers, and inline annotations (\S\ref{sec:tokenizer}, \S\ref{subsec:ocr}).
    \item We show that using OMR output as external context for frontier VLMs (\S\ref{sec:tool-use}) establishes a new state of the art on sheet music understanding  (MusiXQA, \citealp{musixqa}; SSMR-Bench, \citealp{ssmr}; \S\ref{subsec:smu}).
\end{itemize}

We will enable reproduction by releasing data and code upon publication.

\section{Methodology}
\label{sec:methods}

\begin{figure}[tb]
    \centering
    \includegraphics[width=\textwidth]{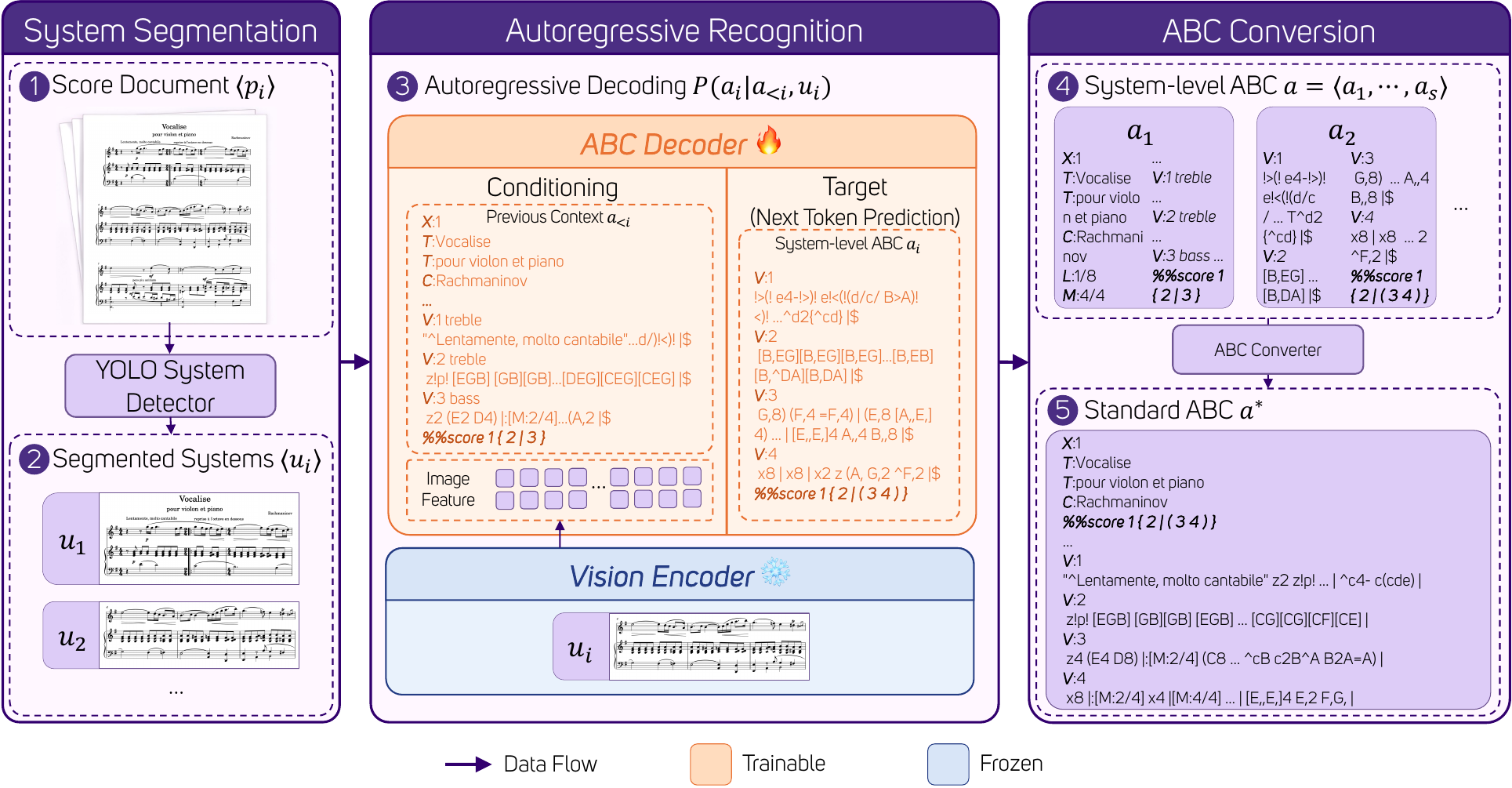}
    \caption{\textbf{Overview of Legato~2 pipeline.} The process consists of three stages: (i) System segmentation~(\S\ref{sec:segmentation}), which uses a YOLO model~\citep{yolov8} to extract musical systems from the document; (ii) Autoregressive recognition~(\S\ref{sec:vlm}), wherein a vision-language model transcribes the current system based on the preceding context and the current image; and (iii) ABC conversion~(\S\ref{sec:system_level_ABC}), which employs a rule-based converter to merge the system-level ABC outputs into a standard ABC format.  %\nascomment{the main issue with the figure is that everything is too small to read.  ideally the system segmentation step would make it clear that this is a relatively simple visual task and maybe also give a sense of how complicated the ABC conversion is.}\guang{tried to make the texts larger, and make room for ABC conversion from the system segmentation box} 
    \label{fig:overview}}
\end{figure}

%\nascomment{we want the reader to feel like we've made a sensible, domain-informed choice about how to break the problem up ... it's not a messy pipeline defined by what was easy to do with available data}

%Our starting point is the state-of-the-art OMR system, Legato~\citep{legato1}, and its pretraining dataset, PDMX-Synth, a large dataset of real digital sheet music~\citep{pdmx} paired with images synthesized to train Legato. 
%Unlike Legato and other previous competitive approaches~\citep{smtpp}, 
%where entire pages of sheet music are concatenated and fed directly into the VLM for full recognition as undifferentiated images,
%we start by segmenting the sheet music into systems (musical lines) using a computer vision model trained on piano sheet music (\S\ref{sec:segmentation}).  The main neural model is trained to map a single segment at a time into musical notation, conditioning on the output from preceding lines (\S\ref{sec:vlm}).
%A final stage merges the system-level outputs into a standard format (ABC; \S\ref{sec:system_level_ABC}).
%We also investigate methods to build a text-aware tokenizer so that our model can recognize embedded textual elements~(\S\ref{sec:tokenizer}).
%Figure~\ref{fig:overview} summarizes the pipeline. 
Our starting point is Legato~1~\citep{legato1}, the previous state-of-the-art OMR model, and its PDMX-Synth training setup. We build on this paradigm but make several targeted departures motivated by the structure of sheet music. First, we segment pages into musical systems rather than treating the page as a single undifferentiated recognition unit (\S\ref{sec:segmentation}). Second, we train the VLM to recognize one system at a time while conditioning on previous system outputs (\S\ref{sec:vlm}). Third, we introduce system-level ABC, an intermediate representation aligned to musical systems and convertible back to standard ABC (\S\ref{sec:system_level_ABC}). Fourth, we make the tokenizer text-aware with byte fallback, allowing the model to preserve embedded textual content rather than replacing it with a placeholder (\S\ref{sec:tokenizer}). Finally, we use the resulting symbolic transcription as context for downstream sheet-music understanding (\S\ref{sec:tool-use}). Figure~\ref{fig:overview} summarizes the pipeline. 
%The proposed system differs from Legato through a set of targeted changes: system detection, system-level autoregressive recognition, system-level ABC conversion, and text-aware tokenization. Our ablations focus on the changes that can be most directly isolated under the Legato architecture: system-level recognition and byte fallback. We run additional ablations that scale tokenizer vocabulary size and vary model architecture, which help determine effective configurations for the proposed pipeline. 

\subsection{System Segmentation}
\label{sec:segmentation}

Similar to previous research~\citep{umust}, we employ the YOLOv8 medium~\citep{yolov8} model ($\sim$26M parameters) to segment systems from each page. The checkpoint released by \citet{umust} was trained exclusively on piano sheet music. To generalize its capability, we finetune using 1,024 manually annotated, rendered pages of sheet music, encompassing layouts ranging from single-staff systems to complex orchestral arrangements. To make the YOLO model more robust, half of the dataset (512 pages) is randomly sampled from the PDMX-Synth training split while the other half comes from IMSLP~\citep{imslp}, a digital library of realistic typeset sheet music images. We provide the training details of this YOLO model in Appendix~\ref{ap:training_detail} and quantify its performance in Appendix~\ref{subsec:yolo_segmentation}.

Formally, given a sheet of music represented as a set of page images $\langle p_1, \ldots, p_n\rangle$, we apply the YOLO model $M_\mathrm{YOLO}$ to each $p_i$ to obtain a sequence of $s_i$ system bounding boxes: $\langle b_{i,1}, \ldots, b_{i,s_i}\rangle = M_\mathrm{YOLO}(p_i)$.  
We align the top edge of $b_{i,1}$ with the top of the page, and the bottom edge of $b_{i,s_i}$ with the bottom of the page.
Combining results from all pages, we get a list of segmented system images defined by the system bounding boxes; we denote these images by $\langle u_1, \ldots, u_s\rangle$.

\subsection{Autoregressive Recognition}
\label{sec:vlm}

%\nascomment{moved earlier so the text follows the pipeline}
%\guang{We introduced $a_i$ and system-level ABC in later sections. Would that be a problem?} \nascomment{good catch.  added brief notation explanation}
The VLM component of Legato~2 is designed to recognize an image of a musical system conditioned on the system-level ABC notation of preceding systems. Specifically, our vision-language model $M_{\mathrm{VLM}}$ is trained to model the distribution over system-level ABC expression sequences  $\langle a_1, \ldots, a_s\rangle$ (explained further in \S\ref{sec:system_level_ABC}) given input images of systems $\langle u_1, \ldots, u_s\rangle$: 
\[
    P(a_1, a_2, \cdots, a_s \mid  u_1, u_2, \cdots, u_s) = \prod_{i} P(a_i \mid a_{<i}, u_1, u_2, \cdots, u_s).
\]
Because the system-level representation of the $i$th system depends only on its corresponding visual representation and preceding context, the right-side factors simplify to $ P(a_i \mid a_{<i}, u_i)$.
Therefore, $M_{\mathrm{VLM}}$ is trained autoregressively to model $P(a_i \mid a_{<i}, u_i)$. 
%\vcomment{maybe i was reading too fast but have you told me what u is I assume its the resulting abc?}\guang{It's introduced at the end of section 3.1. The segmented system-level images.} \nascomment{I changed that slightly, hopefully better now}
The detailed context management strategy for the model is provided in Appendix~\ref{ap:context_management}. 
In our implementation, $a_{<i}$ is left truncated to 1,024 tokens.  Legato~2 has a maximum context length of 2,048 tokens; the other 1,024 are given to the target, $a_i$. 
%\vcomment{how does leaving ai truncated to 1024 deal with the probelm of insufficient context length?} \guang{truncation makes the context shorter so it can be fit into the model. should we call it ``insufficient context length''?} \nascomment{reworked}
%\guang{the maximum context length is 2,048. We give 1,024 to $a_{<i}$ and 1,024 to $a_i$ (target). }

For the vision-language model $M_{\mathrm{VLM}}$, we adopt the same architecture as Legato~1~\citep{legato1} due to its demonstrated effectiveness. The total number of trainable parameters is $113.7$M. The details of training hyper-parameters are shown in Appendix~\ref{ap:training_detail}.
Additionally, we conducted an ablation study to investigate the impact of scaling up the architecture (see \S\ref{sec:ablation}). We also quantify robustness of the VLM against upstream YOLO errors in Appendix~\ref{subsec:yolo_robustness}. %The results indicate that increasing the model size provides marginal benefits under the current data constraints. \vcomment{Potentially move discussion of the ablation to a footnote if it's not essential to understanding?} \nascomment{I would keep mention of the study but don't give the conclusion away here, similar to the tokenization thing earlier}

\subsection{System-Level ABC}
\label{sec:system_level_ABC}

\begin{figure}[tb]
    \centering
    \includegraphics[width=\linewidth]{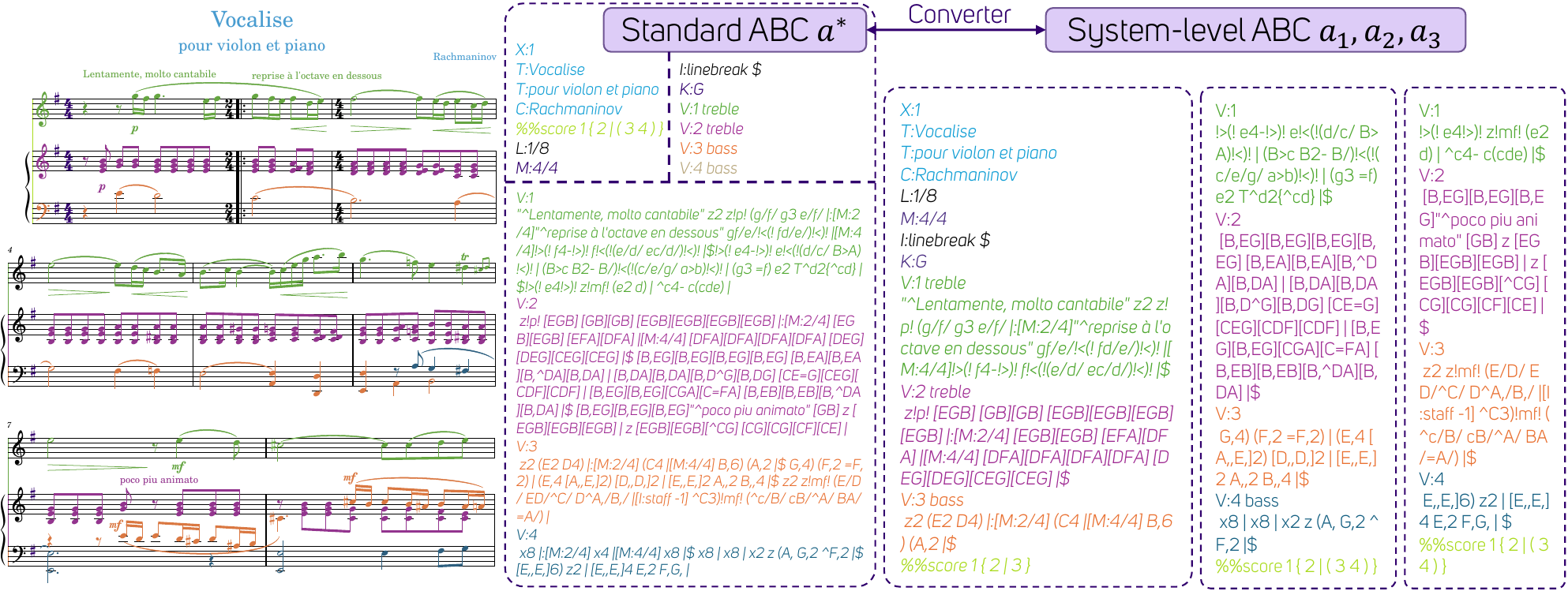}
    \caption{\textbf{An example of a piece of sheet music and its corresponding ABC representations.} Unlike the standard ABC format $a^*$ , our system-level ABC $a = \langle a_1, a_2, a_3\rangle$ is organized system by system, ensuring that the encoding of a given system remains independent of subsequent context. 
    \label{fig:abc}}
\end{figure}

Following \citet{legato1}, we adopt the ABC notation as our target format due to its simplicity and sequential structure. 
The ABC notation in PDMX-Synth is organized in a voice-by-voice sequence, as illustrated in Figures~\ref{fig:overview} and \ref{fig:abc}, which we denote as $a^*$. This representation does not provide a system-level alignment between the systems in the sheet music image and the underlying symbolic music. 
Inspired by SMT-ABC Notation~\citep{mupt}, we introduce a variant of the ABC notation,  ``System-Level ABC''. Formally, the system-level ABC representation of a piece of sheet music with $s$ systems is defined as $a = \langle a_1, \ldots, a_s\rangle$, where $a_i$ denotes the system-level representation of the $i$th system. 
This variant is organized in a system-by-system order. Consequently, the ABC representation $a_i$ for each system depends on its corresponding visual representation and preceding context, but remains otherwise independent of subsequent context; for example, newly introduced voices do not appear in prior systems.
To bridge these formats, we implement a rule-based converter to map between $a^*$ and $a$ (See details in Appendix~\ref{ap:abc_converter}). This enables the ground-truth data in PDMX-Synth to be split into individual systems for training (converting $a^*$ to $a$), while also allowing the VLM's predictions for individual systems to be reconstructed back into the standard ABC format $a^*$.

\subsection{Text-Aware ABC Tokenizer}
\label{sec:tokenizer}

Unlike Legato~1 which replaces all text spans with a single \texttt{<|text|>} token \citep{legato1}, a central goal of our system is to transcribe both musical notation and embedded text, including titles, composer names, staff labels, and inline annotations. Although the BPE over ABC tokenizer introduced in Legato~1 preserves frequent musical patterns, the lack of textual elements in its training data prevents the recovery of the original text. We therefore train a text-aware ABC tokenizer that preserves textual content, while still retaining the ability to represent complex musical concepts efficiently. 

Following PDMX-Synth, we construct the tokenizer-training corpus from PDMX~\citep{pdmx}, but, unlike Legato, we retain all text.\footnote{We continue to exclude lyrics from our corpus as they are both uncommon in PDMX and less tightly coupled to the musical structure than other embedded textual elements; we expect that existing OCR tools can be adapted for lyric reconstruction in future work.} This text-inclusive PDMX-Synth corpus contains relatively sparse text that is dominated by a few common languages, and we find that naive BPE over this corpus allocates vocabulary to many rare characters that are not learned reliably, degrading performance~(Appendix~\ref{ap:naive_tokenizer}). 
To retain the benefits of BPE for musical notation while still supporting arbitrary text, we instead use BPE with byte fallback. We initialize the tokenizer with the 243 valid byte values used in UTF-8 and learn merge rules from the text-inclusive PDMX-Synth corpus. This allows common musical and textual patterns to be represented compactly, while unseen or rare characters can still be represented through their byte sequences. \S\ref{sec:ablation} evaluates the performance of our byte-fallback tokenizer at different vocabulary sizes; we find best performance at a vocabulary size of 4096.

\subsection{OMR for Sheet Music Understanding}
\label{sec:tool-use}

As shown by \cite{legato1}, existing frontier vision-language models fail at sheet music recognition; these results suggest that OMR models can serve as a complement to their capability and eventually improve sheet music understanding. Given an image of sheet music, we first transcribe it with Legato~2, and then provide the resulting transcription to the frontier VLM alongside the original image when prompting it on a downstream task. In this setup, the frontier VLM can optionally use the transcription in addition to the visual input, allowing it to condition on a symbolic representation of the sheet music. We hypothesize that access to this transcription improves the model's music understanding abilities. 
The exact prompts used in this investigation are given in the Appendix~\ref{section:system_prompts}. We test GPT 5~\citep{gpt5} and Gemini 3.1 Pro~\citep{gemini3}, two of the top-performing vision language models, as part of our experiments. 
%\vcomment{did you test asking the questions and giving legato as a tool for the model to use internally? basically seeing if the model chooses to use OMR? not sure if it would help, but curious if you tested with that.}\guang{we didn't test that. I think we directly feed the output of OMR into the model.} \brian{Can we actually call this a tool?} \nascomment{yeah we should be careful here.  I think some readers may find it misleading to call this tool use.  maybe better to say something like ``we evaluate the benefits of augmenting a prompt with our model's output, alongside the image, to the music understanding performance of a VLM'' ... then say something in the discussion section about how ``these findings show the potential of OMR as a tool''}
%\guang{reworked to avoid ``tool-use'' and ``tool call''. I say something like ``context provider'' instead.}

\section{Implementation and Evaluation Details}

\subsection{Baselines}

We evaluate our approach against three distinct baselines. First, \textbf{Legato~1}~\citep{legato1} represents the state of the art in end-to-end neural OMR, utilizing a vision-language architecture to autoregressively transcribe full-page or concatenated multi-page images directly into ABC notation. Second, we compare against \textbf{Audiveris}~\citep{audiveris}, a comprehensive, open-source rule-based system that outputs MusicXML, noting it as the sole prior system equipped to transcribe embedded textual metadata. Finally, we benchmark general-purpose \textbf{Frontier VLMs}---specifically GPT-5 and Gemini 3.1 Pro~\citep{gemini3, gpt5}---on both sheet music recognition and understanding tasks, using the prompts detailed in Appendix~\ref{section:system_prompts}. Note that Audiveris requires input preprocessing, which we detail in Appendix~\ref{section:audiveris_preprocess}.

\subsection{Evaluation Metrics} 

For our main results, we report the Optical Music Recognition - Normalized Edit Distance (OMR-NED)~\citep{omrned}. This is consistent with prior work such as Legato, and offers several benefits over traditional metrics such as Symbol Error Rate \citep{legato1}. More detail is provided in Appendix~\ref{subsec:omr_ned_eval_metric}.

%To evaluate the recognition of embedded text, we use two different metrics depending on models' output.
%First, for comparison with OMR models (e.g. Audiveris) that can output symbolic music, we convert all outputs to ABC notation, and extract the title, composer, and staff names from the header, and then extract any text within the music itself --- this is easily distinguished in ABC as anything within quotes. We clean the text of the location indicators and concatenate in canonical order, using double underscores as text separators. We then compute the character error rate using Levenstein distance.
%For comparison with OCR models that only output bounding boxes with text content, the output cannot be structured in a canonical order, so we compute the set edit distance rather than the character edit distance, using Levenstein distance as cost function for editing one text box into another. We denote this metric as unordered character error rate~(Unordered-CER).

To evaluate embedded text recognition, we use character error rate (CER) for OMR-style symbolic outputs and unordered character error rate (Unordered-CER) for OCR-style bounding-box outputs. For OMR models, we convert outputs to ABC, extract title, composer, and inline quoted text, canonicalize the extracted text, and compute character-level Levenshtein distance. For OCR models, whose bounding boxes have no reliable canonical order, we instead compute a set edit distance over text boxes using Levenshtein distance as the box-level edit cost. Full details are provided in Appendix~\ref{subsec:text_eval_metric}.

\subsection{Evaluation Datasets}

\subsubsection{Sheet Music Recognition}

We employ a series of realistic datasets introduced by Legato~1~\cite{legato1} ---including the test split of PDMX-Synth, OpenScore String Quartets, OpenScore Lieder, and IMSLP Piano Scores---to assess model performance in sheet music recognition. Note that the OpenScore String Quartets and Lieder datasets used in this suite are small evaluation subsets derived from their respective full collections~\citep{OpenScoreLieder, OpenScoreStringQuartets}. For these subsets, the authors of Legato~1 retrieved the corresponding photocopies from IMSLP, providing both software-rendered images and scans of published physical copies. IMSLP Piano Scores is an additional dataset sourced from IMSLP and annotated by the authors of Legato. %\nascomment{this last sentence feels redundant:} In summary, we evaluate our model's performance on the test split of PDMX-Synth, the rendered and camera-captured versions of the OpenScore String Quartets and Lieder datasets, and IMSLP Piano Scores.

\subsubsection{Sheet Music Understanding}

%\paragraph{MusiXQA} We evaluate the performance of our model on sheet music understanding using MusiXQA~\citep{musixqa}. MusiXQA is a visual question-answering benchmark for music notation designed to test whether multimodal language models can read and reason over sheet music. It consists of synthetically generated music pages paired with structured annotations and question-answer tasks covering core elements of written music, including pitch, duration, chords, clefs, key and time signatures, and textual markings. This makes it highly suitable for studying how well vision-language models interpret the semantic content of musical notation. We follow \citet{musixqa} in using G-Acc as the metric for evaluating the accuracy of open-ended responses. Specifically, given the ground-truth reference, a large language model is prompted to judge whether a model’s predicted answer is semantically correct. We test set of MusiXQA is split into two: the simple split and the harder OMR-based split, which requires the model to possess music understanding and OMR abilities. We evaluate on the OMR split because the simple split is too easy for frontier VLMs (baseline methods for both models scored $>90\%$). We use \lm{GPT-5} as the evaluator, as the originally used \lm{GPT-4o} model is no longer publicly accessible.
\paragraph{MusiXQA} We evaluate sheet music understanding on MusiXQA~\citep{musixqa}, a visual question-answering benchmark that tests multimodal reasoning across core musical elements (e.g., pitch, duration, chords, clefs, signatures, and textual markings). The benchmark provides two test splits; we evaluate exclusively on the harder ``OMR split''---which necessitates joint recognition and semantic comprehension---because the ``simple split'' is already saturated by frontier VLMs (scoring $>90\%$). Following \citet{musixqa}, we measure open-ended accuracy using the G-Acc metric, wherein a language model judges the semantic correctness of predictions against a ground-truth reference. Because the originally specified \lm{GPT-4o} is no longer publicly accessible, we substitute \lm{GPT-5} as our evaluator.

\paragraph{SSMR-Bench} We additionally evaluate Legato~2 on the SSMR-Bench dataset~\citep{ssmr}. Unlike MusiXQA, SSMR-Bench is structured as a multiple-choice visual question-answering task, where sheet music images may appear in both the prompts and the candidate choices. The questions focus on sheet music reasoning and are programmatically generated around core musical concepts, including rhythm, chords, intervals, and scales. SSMR-Bench provides both training and evaluation splits, but we only utilize the evaluation set. Following the original methodology, we employ standard accuracy as our evaluation metric.

%\paragraph{SSMR-Bench} To further verify our results, we also evaluate our method on SSMR Bench \cite{ssmr}. SSMR-Bench is a synthetic benchmark for sheet music reasoning, designed to evaluate whether language and multimodal models can interpret musical notation and apply music-theoretic rules rather than merely recognize symbols. It is built from programmatically generated question-answer pairs derived from core concepts such as rhythm, chords, intervals, and scales, and is provided in both textual form using ABC notation and visual form using rendered staff notation. More broadly, SSMR-Bench serves as both an evaluation set and a source of training data for developing models that can reason over sheet music in a structured, verifiable way. We evaluate our method based on raw multiple-choice accuracy. For SSMR Bench, we evaluate \lm{GPT-5} and \lm{Gemini-3-Pro} \citep{gemini3, gpt5}

\subsection{Implementation Details}

For VLM training, we adopt the protocol established in \cite{legato1} (See Appendix~\ref{ap:training_detail}).
Because our VLM is trained exclusively on synthetic sheet music but evaluated on real-world data, optimizing hyperparameters on the PDMX-Synth validation set may cause the model to overfit to synthetic artifacts. To address this, we compile a new validation set comprising 130 pages from the OpenScore String Quartets and OpenScore Lieder datasets, ensuring these pages are strictly excluded from all test sets. For each model checkpoint, we perform a grid search on this validation set over a repetition penalty of $\{1.0,1.1,1.2\}$ and a beam size of $\{1,2,5,10\}$. Finally, we select the checkpoint and its corresponding inference parameters that achieve the lowest OMR-NED on this validation set. See Appendix~\ref{ap:validation} for detailed validation results of all model variants.

\section{Experimental Results}

\subsection{Page-level Recognition}
\label{subsec:page-level}

We evaluate Legato~2 under the exact same setting as Legato; specifically, we provide the model with a single page of sheet music and evaluate its predictions against the ground truth of the corresponding page. 
As shown in Table~\ref{tab:main_result}, Legato~2 outperforms Legato, the previous state of the art, across all datasets. 
As noted by \citet{legato1}, Legato~1 exhibits degraded performance on the OpenScore String Quartets dataset due to the dense visual nature of string quartets, where a single page contains numerous staves of individual voices. 
In contrast, Legato~2 demonstrates significant improvements on these datasets primarily because it operates at the system level. By first segmenting the sheet music into systems, the VLM is able to process each system at a higher effective resolution.

\begin{table}[tb]
\caption{\textbf{Evaluation of page-level sheet music recognition.} Legato~2 consistently outperforms other models across all datasets.}
\label{tab:main_result}
\centering
\begin{tabular}{lcccc}
\toprule 
\multirow{2}{*}{Dataset (\# of Pages)}  & \multicolumn{4}{c}{OMR-NED $\downarrow$} \\
& Audiveris & Gemini 3.1 Pro & Legato~1 & Legato~2 \\
\midrule
{\em PDMX-Synth Test Set} (411)        & $56.3$ & $90.3$ & $28.6$ & $\mathbf{23.5}$ \\
{\em Rendered OpenScore String Quartets} (252) & $64.6$ & $93.5$ & $32.9$ & $\mathbf{17.1}$ \\
{\em Camera OpenScore String Quartets} (252) & $75.4$ & $94.1$ & $58.2$ & $\mathbf{31.6}$ \\
{\em Rendered OpenScore Lieder} (64)         & $76.9$ & $90.9$ & $39.5$ & $\mathbf{27.6}$ \\
{\em Camera OpenScore Lieder} (64)           & $85.8$ & $91.7$ & $44.9$ & $\mathbf{43.6}$ \\
{\em IMSLP Piano Scores (32)}            & $71.5$ & $89.1$ & $44.3$ & $\mathbf{34.2}$ \\
\bottomrule
\end{tabular}
\end{table}

\subsection{Multi-page Recognition}
\label{subsec:multi-page}

%Page-level evaluation provides a reasonable baseline for assessing an OMR model's capabilities. However, in a more realistic setting, users typically input full documents consisting of multiple page images. Previous models approach this task either by concatenating the pages into a single extended image, which can lead to performance degradation, or by processing each page independently and merging the results. The latter approach causes the model to lose cross-page context and often requires complex merging mechanism. In contrast, our model naturally accommodates documents of arbitrary length because the VLM processes musical systems autoregressively. To demonstrate this capability, we evaluate our model on the task of multi-page sheet music recognition.
While page-level evaluation establishes a baseline, real-world applications require multi-page processing. Previous methods either concatenate images (degrading performance) or process pages independently (losing cross-page context and requiring complex merging logic). In contrast, our autoregressive, system-level processing naturally accommodates arbitrary-length documents. We demonstrate this advantage through multi-page sheet music recognition evaluations.
%To this end, we employ the complete OpenScore Lieder dataset, which comprises sheet music ranging from single-page examples to 37 page pieces.
%This comprehensive set is distinct from the 64-page evaluation subset referenced in Table~\ref{tab:main_result}. 
%To build this dataset, we render the document images directly from the source MusicXML files. The resulting samples are partitioned into five categories based on their cumulative aspect ratios. Documents with an aspect ratio greater than or equal to 12 are allocated to a dedicated bin, while the remaining samples are divided among four bins representing equal intervals of aspect ratio values. Because the natural distribution of sheet music across these intervals is uneven, we uniformly sample 100 instances from each bin to ensure a balanced evaluation. Therefore, the total number of documents being evaluated is 500.
To this end, we utilize the full OpenScore Lieder dataset---distinct from the 64-page evaluation subset in Table~\ref{tab:main_result}---which comprises sheet music ranging from 1 to 37 pages per piece. 
We render the document images directly from source MusicXML files and partition them into five bins based on cumulative aspect ratio: one bin for ratios $\ge 12$, and four equally spaced bins for the remainder. 
To counteract naturally uneven length distributions, we uniformly sample 100 instances per bin, yielding a balanced evaluation set of 500 documents.

\begin{figure}[tb]
    \centering
    \includegraphics[width=0.6\textwidth]{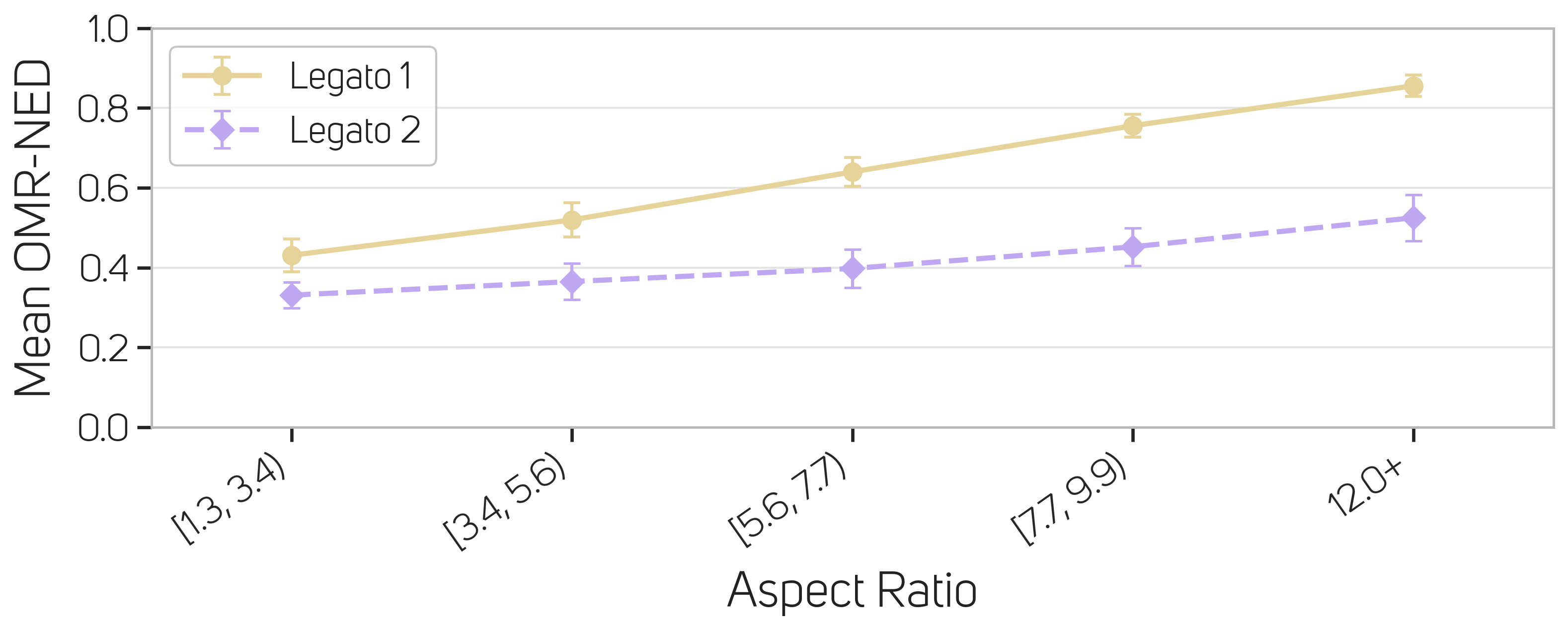}
    \caption{\textbf{Evaluation on multi-page sheet music recognition.} Legato~2 consistently outperforms Legato~1 across all aspect ratio bins, and also degrades at a substantially slower rate. The error bars represent 95\% normal approximated confidence intervals. %\nascomment{can save lots of space by making the figure smaller, shorter, and fatter.  maybe use sidecap package to put caption to its side instead of below.}\brian{I made it fatter but I am not sure if side captions are allowed...}
    }
    \label{fig:multi_page}
\end{figure}

Due to the high computational cost associated with processing large, multi-page images, this analysis is restricted to a direct comparison between Legato~1 and 2. 
As illustrated in Figure~\ref{fig:multi_page}, Legato~2 not only outperforms Legato~1 across all aspect ratio bins, but its performance also degrades at a substantially slower rate, demonstrating robust capabilities across varying document lengths. 
We attribute this stability to our context management strategy: by applying left-truncation to the context window during training, the model naturally adapts to the identical truncation applied during inference. 
This consistency between the training and inference phases effectively mitigates the performance degradation typically associated with exceptionally long documents.

\subsection{Text Recognition}
\label{subsec:ocr}

The text-aware tokenizer in Legato~2 enables neural OMR to recognize embedded textual elements for the first time.
We evaluate this capability on the Camera OpenScore String Quartets dataset used by \cite{legato1}.  We compare against prior text-capable OMR systems and a general-purpose OCR system applied directly to full-page sheet music.  Because Legato~1 and earlier neural OMR models do not process text, we use Audiveris and Gemini 3.1 Pro as representative rule-based and frontier VLM baselines.  
 For the OCR baseline, we employ PaddleOCR~\citep{paddleocr}, a widely used general-purpose OCR system.  Tables~\ref{tab:ocr_audiveris} and~\ref{tab:ocr_paddleocr} report results comparing to the OMR and OCR baselines, respectively. Legato~2 outperforms all baselines on textual components, even when text is evaluated independently of the musical notation.

\begin{table}[tb]
    \caption{\textbf{Evaluation on text recognition.} Lower is better for all metrics. Legato~2 is compared against OMR models (Table~\ref{tab:ocr_audiveris}) and OCR models (Table~\ref{tab:ocr_paddleocr}) on Camera OpenScore String Quartets.}
    \label{tab:ocr}
    \begin{subtable}[t]{0.65\textwidth}
        \centering
        \begin{tabular}[t]{c c c c c}
        \toprule
            \multirow{2}{*}{Model} & \multicolumn{4}{c}{Character Error Rates~(CER) \% $\downarrow$} \\
            \cmidrule(lr){2-5}
             & title & composer & others & Total\\
        \midrule
            Audiveris & $28.2$ & $154.2$ & $95.2$ & $73.8$\\
            Gemini & $18.7$ & $130.8$ & $76.1$ & $58.1$ \\
            Legato~2 & $\mathbf{10.6}$ & $\mathbf{97.2}$ & $\mathbf{31.0}$ & $\mathbf{24.8}$\\
        \bottomrule
        \end{tabular}
        \caption{\textbf{Character error rates for textual elements.} The categories ``\texttt{title}'', ``\texttt{composer}'', and ``\texttt{others}'' denote the error rates for the title, composer name and other embedded annotations. The error for ``\texttt{composer}'' is high because models usually detect page numbers as composer names.}
        \label{tab:ocr_audiveris}
    \end{subtable}
    \hfill
    \begin{subtable}[t]{0.3\textwidth}
        \centering
        \begin{tabular}[t]{c c}
        \toprule
             \multirow{2}{*}{Model} &  Unordered \\
             & CER \% $\downarrow$ \\
        \midrule
             PaddleOCR & $85.5$ \\ 
             Legato~2 & $\mathbf{25.1}$\\
        \bottomrule
        \end{tabular}
        \caption{\textbf{Unordered character error rates for textual elements.} The baseline OCR model likely struggles to distinguish text from the surrounding musical notation.}
        \label{tab:ocr_paddleocr}
    \end{subtable}
    \vspace{-14pt}
\end{table}

\subsection{Sheet Music Understanding}
\label{subsec:smu}

We evaluate frontier VLMs for sheet music understanding under three contexts: no transcription, Legato's transcription, and Legato~2's transcription. As shown in Table~\ref{tab:smu}, OMR context improves performance on both benchmarks. Legato~2's transcriptions yield further gains over Legato's, likely because they are higher quality. Although current OMR models remain imperfect, these results suggest that frontier VLMs can extract information more effectively from symbolic OMR outputs than from raw sheet music images.

\begin{table}[tb]
    \caption{\textbf{Evaluation on sheet music understanding task.} Higher is better for all metrics. The ``Context'' column specifies which OMR model's output is used as context.}
    \label{tab:smu}
    \begin{subtable}[t]{0.48\textwidth}
        \centering
        \begin{tabular}{lcc}
        \toprule
        Context & Gemini 3 Flash & GPT 5 Mini \\
        \midrule
        None & $\phantom{0}8.4$  & $\phantom{0}8.0$ \\
        Legato~1 & $20.1$  & $16.6$  \\
        Legato~2 & $\mathbf{25.3}$ & $\mathbf{20.2}$ \\
        \bottomrule
        \end{tabular}
        \caption{\textbf{G-Acc $\uparrow$ on MusiXQA.}}
    \end{subtable}
    \hfill
    \begin{subtable}[t]{0.48\textwidth}
        \centering
        \begin{tabular}{lcc}
        \toprule
        Context & Gemini 3.1 Pro & GPT-5 \\
        \midrule
        None  & $71.4$ & $51.8$ \\
        Legato~1   & $84.8$ & $65.3$ \\
        Legato~2 & $\mathbf{92.7}$ & $\mathbf{71.7}$ \\
        \bottomrule
        \end{tabular}
        \caption{\textbf{Accuracy $\uparrow$ on SSMR-Bench.}}
    \end{subtable}
    \vspace{-7pt}
\end{table}

% \begin{table}[tb]
%     \caption{\textbf{Performance on MusiXQA and SSMR Bench.} Providing our model's transcriptions as context to VLMs enables strong performance boosts.}
%     \label{tab:musicqa-ssmr-side-by-side}
%     \begin{minipage}[t]{0.48\textwidth}
%         \centering
%         \textbf{MusiXQA}\\[0.5em]
%         \resizebox{\linewidth}{!}{
%             \begin{tabular}{lcc}
%             \toprule
%             Transcription & Gemini 3 Flash & GPT 5 Mini \\
%             \midrule
%             Baseline & 0.084  & 0.0804 \\
%             Legato~1 & 0.201  & 0.166  \\
%             Our Model & \textbf{0.253} & \textbf{0.202} \\
%             \bottomrule
%             \end{tabular}
%         }
%     \end{minipage}
%     \hfill
%     \begin{minipage}[t]{0.44\textwidth}
%     \centering
%     \textbf{SSMR Bench}\\[0.5em]
%     \resizebox{\linewidth}{!}{
%     \begin{tabular}{lcc}
%     \toprule
%     Transcription & Gemini 3.1 Pro & GPT-5 \\
%     \midrule
%     Baseline  & 0.714 & 0.518 \\
%     Legato    & 0.848 & 0.653 \\
%     Our Model & \textbf{0.927} & \textbf{0.717} \\
%     \bottomrule
%     \end{tabular}
%     }
%     \end{minipage}
% \end{table}

\subsection{Ablation Studies}
\label{sec:ablation}

\paragraph{System Segmentation and Byte Fallback}
We first investigate the effectiveness of two key design choices in Legato~2: system segmentation and byte fallback. Starting from the baseline Legato~1 architecture, we first incorporate the system segmentation mechanism; specifically, we train an identical Legato~1 model from scratch under the system-by-system conditional generation setting. Note that this intermediate model retains the original Legato~1 tokenizer. Next, we replace this tokenizer with the byte-fallback variant described in \S\ref{sec:tokenizer}, and retrain both the tokenizer and the model. To ensure a fair comparison, we maintain a vocabulary size of 2,048, identical to that of the original Legato. The performance of these two variants, alongside the base Legato~1 model, is reported in the left three columns of Table~\ref{tab:ablation}. The results demonstrate that system segmentation improves performance by a large margin. Furthermore, the incorporation of byte fallback, while enabling the model to process embedded textual content, yields comparable performance.

% \begin{table}[tb]
% \caption{\textbf{Ablation Study on System Segmentation (\texttt{SS}) and Byte Fallback (\texttt{BF}).} Starting from Legato, we add system segmentation and byte fallback into the architecture.}
% \label{tab:tokenizer_ablation}
% \centering
% \begin{tabular}{lccc}
% \toprule
%  \multirow{2}{*}{Dataset (\# of Pages)}& \multicolumn{3}{c}{OMR-NED $\downarrow$} \\
%  & Legato & \texttt{+SS} & \texttt{+SS +BF} \\
% \midrule 
% {\em Validation Set} (130) & $70.2$  & $\mathbf{48.5}$ & $\mathbf{48.5}$ \\
% \midrule
% {\em PDMX-Synth Test Set} (411)          & $28.6$ & $\mathbf{25.8}$ & $26.4$ \\
% {\em Rendered OpenScore String Quartets} (252) & $32.9$ & $17.9$ & $\mathbf{16.4}$ \\
% {\em Camera OpenScore String Quartets} (252)   & $58.2$ & $\mathbf{35.2}$ & $35.6$ \\
% {\em Rendered OpenScore Lieder} (64)           & $39.5$ & $29.1$ & $\mathbf{28.7}$ \\
% {\em Camera OpenScore Lieder} (64)             & $44.9$ & $45.8$ & $\mathbf{44.8}$ \\
% {\em IMSLP Piano Scores} (32)              & $44.3$ & $\mathbf{34.3}$ & $35.1$ \\
% \bottomrule
% \end{tabular}
% \end{table}

\paragraph{Vocabulary Size}
Based on the previous model (Legato~1 \texttt{+SS+BF}), we then investigate the effect of different vocabulary sizes on Legato~2's performance. 
Maintaining the tokenizer construction methodology detailed in \S\ref{sec:tokenizer}, we vary only the final vocabulary size. 
While larger vocabulary sizes enable the tokenizer to capture more common patterns from the training data, they also increase the risk of overfitting. 
Right three columns of Table~\ref{tab:ablation} present the validation and test errors across the evaluated vocabulary sizes $\{2048, 4096, 8192\}$.
Instead of Legato's vocabulary size of 2048, we selected a vocabulary size of 4,096 strictly based on achieving the lowest validation error, rather than test error. %\nascomment{maybe add a point about Legato~1's vocab size of 2048, making this choice a difference relative to that.  in general, I think we need to bring out (1) what exactly is different from Legato~1, and (2) what we learn from making those changes.  section 4. is helpful but it's a bit buried.  need to give the reader more confidence earlier on that our experiments are carefully controlled and not changing tons of things all at once} \brian{partially addressed}

\begin{table}[tb]
\caption{\textbf{Ablation study on system segmentation (\texttt{SS}), byte fallback (\texttt{BF}), and vocabulary size.} Starting from the Legato~1 baseline, we incorporate system segmentation and byte fallback into the architecture, and subsequently increase the vocabulary size from 2,048 to 4,096 (\texttt{+V4096}) and 8,192 (\texttt{+V8192}). The modifications from left to right are cumulative. $\dagger$: Indicates our final model, Legato~2, selected based on validation error.}
\label{tab:ablation}
\centering
\begin{tabular}{lccccc}
\toprule
 \multirow{2}{*}{Dataset (\# of Pages)}& \multicolumn{5}{c}{OMR-NED $\downarrow$} \\
 & Legato~1 & \texttt{+SS} & \texttt{+BF} & \texttt{+V4096}$^\dagger$ & \texttt{+V8192} \\
\midrule 
{\em Validation Set} (130) & $70.2$  & $48.5$ & $48.5$ & $\mathbf{46.8}$ &$52.8$\\
\midrule
{\em PDMX-Synth Test Set} (411)          & $28.6$ & $25.8$ & $26.4$ & $\mathbf{23.5}$ & $\mathbf{23.5}$ \\
{\em Rendered OpenScore String Quartets} (252) & $32.9$ & $17.9$ & $\mathbf{16.4}$ & $17.1$ & $16.6$ \\
{\em Camera OpenScore String Quartets} (252)   & $58.2$ & $35.2$ & $35.6$ & $\mathbf{31.6}$ & $33.5$\\
{\em Rendered OpenScore Lieder} (64)           & $39.5$ & $29.1$ & $28.7$ & $\mathbf{27.6}$ & $31.3$ \\
{\em Camera OpenScore Lieder} (64)             & $44.9$ & $45.8$ & $44.8$ & $43.6$ & $\mathbf{42.7}$\\
{\em IMSLP Piano Scores} (32)              & $44.3$ & $34.3$ & $35.1$ & $\mathbf{34.2}$ & $37.8$ \\
\bottomrule
\end{tabular}
\end{table}

%\paragraph{Additional Ablation Studies}
%In addition, we explore more potential modifications to improve our model architecture~(Appendix \ref{ap:model_arcitecture}), and find out that these modification only help minimally. We hypothesize that the distribution gap between training and evaluation datasets hinders the performance gains from scaled models, and show that in Appendix \ref{ap:data_distribution}.
%Besides, we also investigate the model with a BPE tokenizer naively trained on text-inclusive PDMX-Synth dataset (Appendix \ref{ap:naive_tokenizer}).
\paragraph{Additional Ablation Studies}
We explore further architectural modifications, which ultimately yield minimal improvements (Appendix~\ref{ap:model_arcitecture}). As detailed in Appendix~\ref{ap:data_distribution}, we hypothesize these limited scaling benefits stem from the distribution gap between the training and evaluation datasets. Finally, we investigate a variant employing a BPE tokenizer naively trained on the text-inclusive PDMX-Synth dataset (Appendix~\ref{ap:naive_tokenizer}).

\subsection{Qualitative Results}

\begin{figure}[tb]
    \centering
    \includegraphics[width=0.8\linewidth]{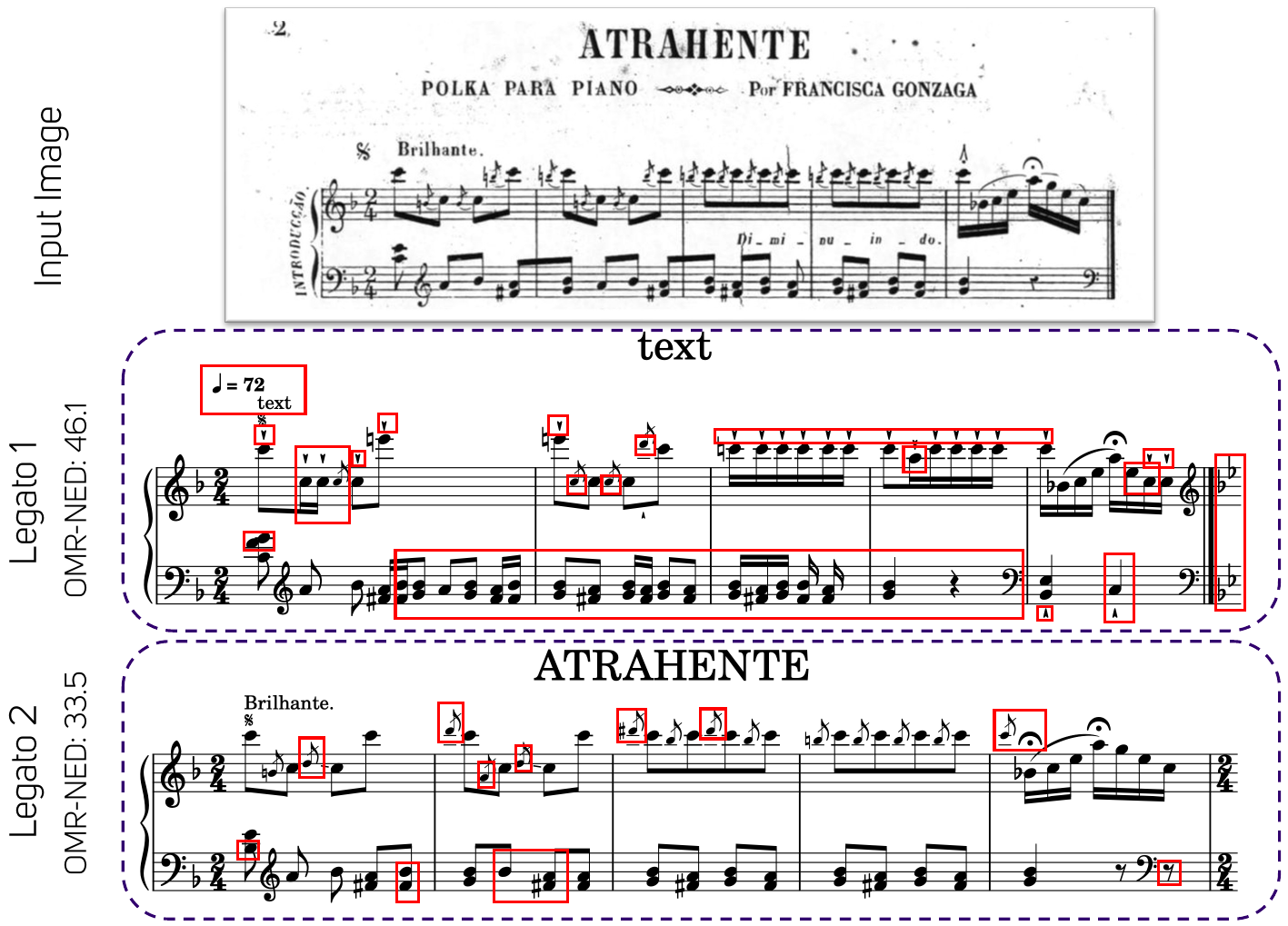}
    \caption{\textbf{Qualitative example from the IMSLP Piano Scores dataset.} The red boxes represent errors from the ground truth. This example was selected because the OMR-NED scores for both models closely approximate their respective dataset averages. Although only the first system is shown here, the models process the full-page image as input. Note that the input's lower staff exhibits a rare formatting irregularity where beams cross the barlines.}
    \label{fig:omr_example}
\end{figure}

\paragraph{Sheet Music Recognition}
Figure~\ref{fig:omr_example} illustrates an example from the IMSLP Piano Scores dataset. 
This example was selected because the OMR-NED scores for both models closely approximate their respective averages across the entire dataset. 
For this specific example, Legato~1 and Legato~2 achieve OMR-NED scores of $46.1$ and $33.5$, respectively. 
As observed, both models fail to transcribe the subtitle and the ``\texttt{Diminuindo.}'' annotation, an omission likely attributable to the unconventional placement of these textual elements. Furthermore, while Legato~1 only detects the spatial position of the title, Legato~2 accurately transcribes the exact text. % \nascomment{maybe drop the ``despite'' since we haven't said anything about multilinguality before}. \brian{@Guang I am unsure about this one, is most of our PDMX data in English?}
At the beginning of the sheet music, Legato~1 generates a redundant tempo mark; this is likely because the model is confused by the ``\texttt{Brilhante.}'' annotation. 
In contrast, Legato~2 processes this content correctly. 
Notably, Legato~1 also misses most of the grace notes, misclassifying them as staccatissimo marks. 
Conversely, because Legato~2 operates on a system-by-system basis, it captures these fine-grained visual details much more effectively. 
Finally, it is worth noting that the input sheet music contains formatting irregularities; specifically, on the lower staff, several beams cross the barlines. 
These visual artifacts disrupt Legato~1's predictions, causing it to generate an output that improperly mixes these measures. 
Although Legato~2 generates redundant chords in the first two measures, it quickly recovers to produce correct predictions in measures 3 and 4, demonstrating  greater robustness to such layout anomalies.

\paragraph{Sheet Music Understanding} We also provide a sheet music understanding example in Appendix~\ref{subsec:sheet_music_understanding}, where we show that \lm{Gemini-3-Flash}~\citep{gemini3flash} successfully uses and cites Legato~2's transcription when responding to a question from SSMR-Bench. 

\section{Conclusions and Future Work}

%\brian{Required by the neurips checklist}
%These two paragraphs talk about limitation and future work, required by neurips
%Our work suggests a broader role for OMR as a tool for sheet music understanding. Rather than treating transcription as an end in itself, we show that symbolic OMR outputs can provide useful musical context for general-purpose VLMs, enabling them to reason more effectively about domain-specific tasks. This points toward future modular systems in which specialized music tools, including OMR, transcription, and analysis models, complement general VLMs on complex musical problems.

%At the same time, our approach has several limitations. First, we were not able to run extensive ablations on individual parts of the pipeline due to resource constraints. In particular, future work should more directly study the system's robustness to errors from the YOLO-based detector and measure how segmentation or recognition errors affect downstream reasoning when OMR output is used as context. Second, our approach is limited by data-distribution differences between the training data, our benchmarks, and real-world sheet music. Scores vary widely in engraving style, repertoire, scan quality, camera capture conditions, historical notation, handwritten markings, and embedded text, and these shifts may reduce both transcription accuracy and the usefulness of the resulting OMR context. Future work should therefore broaden the training and evaluation distributions and develop diagnostic benchmarks that test when OMR can reliably serve as a tool for general music understanding.

In this paper, we introduced Legato~2, an optical music recognition pipeline that segments sheet music into individual systems and trains a vision-language model to recognize each system autoregressively.
We demonstrated that this approach achieves state-of-the-art performance in both single-page and multi-page sheet music recognition. Furthermore, we integrated the capability to recognize embedded textual elements into Legato~2, surpassing previous OMR baselines. 
Finally, we investigated the potential of using OMR as an upstream context provider for broader sheet music understanding, demonstrating that specialized OMR models can effectively complement general-purpose large vision-language models on domain-specific tasks. Looking forward, we envision a framework where diverse, specialized musical tools—including OMR, music transcription, and analysis software—can be used by general VLMs to solve complex, open-ended musical problems. Such a modular architecture promises greater flexibility, efficiency, and accuracy compared to monolithic general-purpose models.

%\begin{ack}
%placeholder
%\end{ack}

\bibliographystyle{plainnat}
\bibliography{reference}

@article{Aucoin2025ClassicalMusic,
  author  = {Aucoin, Matthew},
  title   = {Do You Actually Know What Classical Music Is? {Does} Anyone?},
  journal = {The Atlantic},
  year    = {2025},
  month   = apr,
  day     = {15},
  url     = {https://www.theatlantic.com/magazine/archive/2025/05/aucoin-what-is-classical-music/682119/}
}

@inproceedings{completeomr,
  author       = {Guang Yang and
                  Muru Zhang and
                  Lin Qiu and
                  Yanming Wan and
                  Noah A. Smith},
  title        = {Toward a More Complete OMR Solution},
  booktitle    = {Proceedings of the 25th International Society for
                   Music Information Retrieval Conference
                  },
  year         = 2024,
  pages        = {930-937},
  publisher    = {ISMIR},
  month        = nov,
  venue        = {San Francisco, California, USA and Online},
  doi          = {10.5281/zenodo.14877483},
  url          = {https://doi.org/10.5281/zenodo.14877483},
}

@misc{legato1,
      title={LEGATO: Large-scale End-to-end Generalizable Approach to Typeset {OMR}}, 
      author={Guang Yang and Victoria Ebert and Nazif Tamer and Brian Siyuan Zheng and Luiza Pozzobon and Noah A. Smith},
      year={2025},
      eprint={2506.19065},
      archivePrefix={arXiv},
      primaryClass={cs.CV},
      url={https://arxiv.org/abs/2506.19065}, 
}

@misc{omrned,
      title={Sheet Music Benchmark: Standardized Optical Music Recognition Evaluation}, 
      author={Juan C. Martinez-Sevilla and Joan Cerveto-Serrano and Noelia Luna and Greg Chapman and Craig Sapp and David Rizo and Jorge Calvo-Zaragoza},
      year={2025},
      eprint={2506.10488},
      archivePrefix={arXiv},
      primaryClass={cs.CV},
      url={https://arxiv.org/abs/2506.10488}, 
}

@misc{smtpp,
      title={End-to-End Full-Page Optical Music Recognition for Pianoform Sheet Music}, 
      author={Antonio Ríos-Vila and Jorge Calvo-Zaragoza and David Rizo and Thierry Paquet},
      year={2025},
      eprint={2405.12105},
      archivePrefix={arXiv},
      primaryClass={cs.CV},
      url={https://arxiv.org/abs/2405.12105}, 
}

@misc{musixqa,
      title={{MusiXQA}: Advancing Visual Music Understanding in Multimodal Large Language Models}, 
      author={Jian Chen and Wenye Ma and Penghang Liu and Wei Wang and Tengwei Song and Ming Li and Chenguang Wang and Jiayu Qin and Ruiyi Zhang and Changyou Chen},
      year={2025},
      eprint={2506.23009},
      archivePrefix={arXiv},
      primaryClass={cs.CV},
      url={https://arxiv.org/abs/2506.23009}, 
}

@misc{gemini3flash,
  title        = {Gemini 3 Flash Model Card},
  author       = {{Google DeepMind}},
  year         = {2026},
  howpublished = {https://storage.googleapis.com/deepmind-media/Model-Cards/Gemini-3-Flash-Model-Card.pdf},
  note         = {Model card for the Gemini 3 Flash multimodal model}
}

@misc{ssmr,
      title={Towards an {AI} Musician: Synthesizing Sheet Music Problems for Musical Reasoning}, 
      author={Zhilin Wang and Zhe Yang and Yun Luo and Yafu Li and Xiaoye Qu and Ziqian Qiao and Haoran Zhang and Runzhe Zhan and Derek F. Wong and Jizhe Zhou and Yu Cheng},
      year={2025},
      eprint={2509.04059},
      archivePrefix={arXiv},
      primaryClass={cs.CL},
      url={https://arxiv.org/abs/2509.04059}, 
}

@TECHREPORT{mllama,
    title={{Llama} 3.2: Revolutionizing edge {AI} and vision with open, customizable models},
    author={AI@Meta},
    institution={Meta Platforms, Inc.},
    month=sep,
    year=2024,
    url={https://ai.meta.com/blog/llama-3-2-connect-2024-vision-edge-mobile-devices/}
}

@misc{bai2025qwen3vltechnicalreport,
      title={{Qwen3-VL} Technical Report}, 
      author={Shuai Bai and Yuxuan Cai and Ruizhe Chen and Keqin Chen and Xionghui Chen and Zesen Cheng and Lianghao Deng and Wei Ding and Chang Gao and Chunjiang Ge and Wenbin Ge and Zhifang Guo and Qidong Huang and Jie Huang and Fei Huang and Binyuan Hui and Shutong Jiang and Zhaohai Li and Mingsheng Li and Mei Li and Kaixin Li and Zicheng Lin and Junyang Lin and Xuejing Liu and Jiawei Liu and Chenglong Liu and Yang Liu and Dayiheng Liu and Shixuan Liu and Dunjie Lu and Ruilin Luo and Chenxu Lv and Rui Men and Lingchen Meng and Xuancheng Ren and Xingzhang Ren and Sibo Song and Yuchong Sun and Jun Tang and Jianhong Tu and Jianqiang Wan and Peng Wang and Pengfei Wang and Qiuyue Wang and Yuxuan Wang and Tianbao Xie and Yiheng Xu and Haiyang Xu and Jin Xu and Zhibo Yang and Mingkun Yang and Jianxin Yang and An Yang and Bowen Yu and Fei Zhang and Hang Zhang and Xi Zhang and Bo Zheng and Humen Zhong and Jingren Zhou and Fan Zhou and Jing Zhou and Yuanzhi Zhu and Ke Zhu},
      year={2025},
      eprint={2511.21631},
      archivePrefix={arXiv},
      primaryClass={cs.CV},
      url={https://arxiv.org/abs/2511.21631}, 
}

@inproceedings{wildscore,
    title = "{W}ild{S}core: Benchmarking {MLLM}s in-the-Wild Symbolic Music Reasoning",
    author = "Mundada, Gagan  and
      Vishe, Yash  and
      Namburi, Amit  and
      Xu, Xin  and
      Novack, Zachary  and
      McAuley, Julian  and
      Wu, Junda",
    editor = "Christodoulopoulos, Christos  and
      Chakraborty, Tanmoy  and
      Rose, Carolyn  and
      Peng, Violet",
    booktitle = "Proceedings of the 2025 Conference on Empirical Methods in Natural Language Processing",
    month = nov,
    year = "2025",
    address = "Suzhou, China",
    publisher = "Association for Computational Linguistics",
    url = "https://aclanthology.org/2025.emnlp-main.853/",
    doi = "10.18653/v1/2025.emnlp-main.853",
    pages = "16847--16863",
    ISBN = "979-8-89176-332-6"
}

@misc{gpt5,
      title={{OpenAI GPT-5} System Card}, 
      author={Aaditya Singh and Adam Fry and Adam Perelman and Adam Tart and Adi Ganesh and Ahmed El-Kishky and Aidan McLaughlin and Aiden Low and AJ Ostrow and Akhila Ananthram and Akshay Nathan and Alan Luo and Alec Helyar and Aleksander Madry and Aleksandr Efremov and Aleksandra Spyra and Alex Baker-Whitcomb and Alex Beutel and Alex Karpenko and Alex Makelov and Alex Neitz and Alex Wei and Alexandra Barr and Alexandre Kirchmeyer and Alexey Ivanov and Alexi Christakis and Alistair Gillespie and Allison Tam and Ally Bennett and Alvin Wan and Alyssa Huang and Amy McDonald Sandjideh and Amy Yang and Ananya Kumar and Andre Saraiva and Andrea Vallone and Andrei Gheorghe and Andres Garcia Garcia and Andrew Braunstein and Andrew Liu and Andrew Schmidt and Andrey Mereskin and Andrey Mishchenko and Andy Applebaum and Andy Rogerson and Ann Rajan and Annie Wei and Anoop Kotha and Anubha Srivastava and Anushree Agrawal and Arun Vijayvergiya and Ashley Tyra and Ashvin Nair and Avi Nayak and Ben Eggers and Bessie Ji and Beth Hoover and Bill Chen and Blair Chen and Boaz Barak and Borys Minaiev and Botao Hao and Bowen Baker and Brad Lightcap and Brandon McKinzie and Brandon Wang and Brendan Quinn and Brian Fioca and Brian Hsu and Brian Yang and Brian Yu and Brian Zhang and Brittany Brenner and Callie Riggins Zetino and Cameron Raymond and Camillo Lugaresi and Carolina Paz and Cary Hudson and Cedric Whitney and Chak Li and Charles Chen and Charlotte Cole and Chelsea Voss and Chen Ding and Chen Shen and Chengdu Huang and Chris Colby and Chris Hallacy and Chris Koch and Chris Lu and Christina Kaplan and Christina Kim and CJ Minott-Henriques and Cliff Frey and Cody Yu and Coley Czarnecki and Colin Reid and Colin Wei and Cory Decareaux and Cristina Scheau and Cyril Zhang and Cyrus Forbes and Da Tang and Dakota Goldberg and Dan Roberts and Dana Palmie and Daniel Kappler and Daniel Levine and Daniel Wright and Dave Leo and David Lin and David Robinson and Declan Grabb and Derek Chen and Derek Lim and Derek Salama and Dibya Bhattacharjee and Dimitris Tsipras and Dinghua Li and Dingli Yu and DJ Strouse and Drew Williams and Dylan Hunn and Ed Bayes and Edwin Arbus and Ekin Akyurek and Elaine Ya Le and Elana Widmann and Eli Yani and Elizabeth Proehl and Enis Sert and Enoch Cheung and Eri Schwartz and Eric Han and Eric Jiang and Eric Mitchell and Eric Sigler and Eric Wallace and Erik Ritter and Erin Kavanaugh and Evan Mays and Evgenii Nikishin and Fangyuan Li and Felipe Petroski Such and Filipe de Avila Belbute Peres and Filippo Raso and Florent Bekerman and Foivos Tsimpourlas and Fotis Chantzis and Francis Song and Francis Zhang and Gaby Raila and Garrett McGrath and Gary Briggs and Gary Yang and Giambattista Parascandolo and Gildas Chabot and Grace Kim and Grace Zhao and Gregory Valiant and Guillaume Leclerc and Hadi Salman and Hanson Wang and Hao Sheng and Haoming Jiang and Haoyu Wang and Haozhun Jin and Harshit Sikchi and Heather Schmidt and Henry Aspegren and Honglin Chen and Huida Qiu and Hunter Lightman and Ian Covert and Ian Kivlichan and Ian Silber and Ian Sohl and Ibrahim Hammoud and Ignasi Clavera and Ikai Lan and Ilge Akkaya and Ilya Kostrikov and Irina Kofman and Isak Etinger and Ishaan Singal and Jackie Hehir and Jacob Huh and Jacqueline Pan and Jake Wilczynski and Jakub Pachocki and James Lee and James Quinn and Jamie Kiros and Janvi Kalra and Jasmyn Samaroo and Jason Wang and Jason Wolfe and Jay Chen and Jay Wang and Jean Harb and Jeffrey Han and Jeffrey Wang and Jennifer Zhao and Jeremy Chen and Jerene Yang and Jerry Tworek and Jesse Chand and Jessica Landon and Jessica Liang and Ji Lin and Jiancheng Liu and Jianfeng Wang and Jie Tang and Jihan Yin and Joanne Jang and Joel Morris and Joey Flynn and Johannes Ferstad and Johannes Heidecke and John Fishbein and John Hallman and Jonah Grant and Jonathan Chien and Jonathan Gordon and Jongsoo Park and Jordan Liss and Jos Kraaijeveld and Joseph Guay and Joseph Mo and Josh Lawson and Josh McGrath and Joshua Vendrow and Joy Jiao and Julian Lee and Julie Steele and Julie Wang and Junhua Mao and Kai Chen and Kai Hayashi and Kai Xiao and Kamyar Salahi and Kan Wu and Karan Sekhri and Karan Sharma and Karan Singhal and Karen Li and Kenny Nguyen and Keren Gu-Lemberg and Kevin King and Kevin Liu and Kevin Stone and Kevin Yu and Kristen Ying and Kristian Georgiev and Kristie Lim and Kushal Tirumala and Kyle Miller and Lama Ahmad and Larry Lv and Laura Clare and Laurance Fauconnet and Lauren Itow and Lauren Yang and Laurentia Romaniuk and Leah Anise and Lee Byron and Leher Pathak and Leon Maksin and Leyan Lo and Leyton Ho and Li Jing and Liang Wu and Liang Xiong and Lien Mamitsuka and Lin Yang and Lindsay McCallum and Lindsey Held and Liz Bourgeois and Logan Engstrom and Lorenz Kuhn and Louis Feuvrier and Lu Zhang and Lucas Switzer and Lukas Kondraciuk and Lukasz Kaiser and Manas Joglekar and Mandeep Singh and Mandip Shah and Manuka Stratta and Marcus Williams and Mark Chen and Mark Sun and Marselus Cayton and Martin Li and Marvin Zhang and Marwan Aljubeh and Matt Nichols and Matthew Haines and Max Schwarzer and Mayank Gupta and Meghan Shah and Melody Huang and Meng Dong and Mengqing Wang and Mia Glaese and Micah Carroll and Michael Lampe and Michael Malek and Michael Sharman and Michael Zhang and Michele Wang and Michelle Pokrass and Mihai Florian and Mikhail Pavlov and Miles Wang and Ming Chen and Mingxuan Wang and Minnia Feng and Mo Bavarian and Molly Lin and Moose Abdool and Mostafa Rohaninejad and Nacho Soto and Natalie Staudacher and Natan LaFontaine and Nathan Marwell and Nelson Liu and Nick Preston and Nick Turley and Nicklas Ansman and Nicole Blades and Nikil Pancha and Nikita Mikhaylin and Niko Felix and Nikunj Handa and Nishant Rai and Nitish Keskar and Noam Brown and Ofir Nachum and Oleg Boiko and Oleg Murk and Olivia Watkins and Oona Gleeson and Pamela Mishkin and Patryk Lesiewicz and Paul Baltescu and Pavel Belov and Peter Zhokhov and Philip Pronin and Phillip Guo and Phoebe Thacker and Qi Liu and Qiming Yuan and Qinghua Liu and Rachel Dias and Rachel Puckett and Rahul Arora and Ravi Teja Mullapudi and Raz Gaon and Reah Miyara and Rennie Song and Rishabh Aggarwal and RJ Marsan and Robel Yemiru and Robert Xiong and Rohan Kshirsagar and Rohan Nuttall and Roman Tsiupa and Ronen Eldan and Rose Wang and Roshan James and Roy Ziv and Rui Shu and Ruslan Nigmatullin and Saachi Jain and Saam Talaie and Sam Altman and Sam Arnesen and Sam Toizer and Sam Toyer and Samuel Miserendino and Sandhini Agarwal and Sarah Yoo and Savannah Heon and Scott Ethersmith and Sean Grove and Sean Taylor and Sebastien Bubeck and Sever Banesiu and Shaokyi Amdo and Shengjia Zhao and Sherwin Wu and Shibani Santurkar and Shiyu Zhao and Shraman Ray Chaudhuri and Shreyas Krishnaswamy and Shuaiqi and Xia and Shuyang Cheng and Shyamal Anadkat and Simón Posada Fishman and Simon Tobin and Siyuan Fu and Somay Jain and Song Mei and Sonya Egoian and Spencer Kim and Spug Golden and SQ Mah and Steph Lin and Stephen Imm and Steve Sharpe and Steve Yadlowsky and Sulman Choudhry and Sungwon Eum and Suvansh Sanjeev and Tabarak Khan and Tal Stramer and Tao Wang and Tao Xin and Tarun Gogineni and Taya Christianson and Ted Sanders and Tejal Patwardhan and Thomas Degry and Thomas Shadwell and Tianfu Fu and Tianshi Gao and Timur Garipov and Tina Sriskandarajah and Toki Sherbakov and Tomer Kaftan and Tomo Hiratsuka and Tongzhou Wang and Tony Song and Tony Zhao and Troy Peterson and Val Kharitonov and Victoria Chernova and Vineet Kosaraju and Vishal Kuo and Vitchyr Pong and Vivek Verma and Vlad Petrov and Wanning Jiang and Weixing Zhang and Wenda Zhou and Wenlei Xie and Wenting Zhan and Wes McCabe and Will DePue and Will Ellsworth and Wulfie Bain and Wyatt Thompson and Xiangning Chen and Xiangyu Qi and Xin Xiang and Xinwei Shi and Yann Dubois and Yaodong Yu and Yara Khakbaz and Yifan Wu and Yilei Qian and Yin Tat Lee and Yinbo Chen and Yizhen Zhang and Yizhong Xiong and Yonglong Tian and Young Cha and Yu Bai and Yu Yang and Yuan Yuan and Yuanzhi Li and Yufeng Zhang and Yuguang Yang and Yujia Jin and Yun Jiang and Yunyun Wang and Yushi Wang and Yutian Liu and Zach Stubenvoll and Zehao Dou and Zheng Wu and Zhigang Wang},
      year={2025},
      eprint={2601.03267},
      archivePrefix={arXiv},
      primaryClass={cs.CL},
      url={https://arxiv.org/abs/2601.03267}, 
}

@ARTICLE{umust,
  author={Jung, Jongmin and Kim, Dongmin and Lee, Sihun and Cho, Seola and So, Hyungjoon and Bukey, Irmak and Donahue, Chris and Jeong, Dasaem},
  journal={IEEE Transactions on Audio, Speech and Language Processing}, 
  title={{U-MusT}: A Unified Framework for Cross-modal Translation of Score Images, Symbolic Music, and Performance Audio}, 
  year={2025},
  pages={1-16},
  doi={10.1109/TASLPRO.2025.3648794}
}

@misc{yolov8,
  author = {Glenn Jocher and Ayush Chaurasia and Jing Qiu},
  title = {Ultralytics {YOLO}},
  version = {8.0.0},
  year = {2023},
  url = {https://github.com/ultralytics/ultralytics},
  orcid = {0000-0001-5950-6979, 0000-0002-7603-6750, 0000-0003-3783-7069},
  license = {AGPL-3.0}
}

@techreport{gemini3,
  author      = {{Google DeepMind}},
  title       = {{Gemini 3.1 Pro} Model Card},
  institution = {Google DeepMind},
  type        = {Model Card},
  year        = {2026},
  month       = feb,
  url         = {https://storage.googleapis.com/deepmind-media/Model-Cards/Gemini-3-1-Pro-Model-Card.pdf},
  note        = {Published February 2026}
}

@INPROCEEDINGS{pdmx,
  author={Long, Phillip and Novack, Zachary and Berg-Kirkpatrick, Taylor and McAuley, Julian},
  booktitle={ICASSP 2025 - 2025 IEEE International Conference on Acoustics, Speech and Signal Processing (ICASSP)}, 
  title={{PDMX}: A Large-Scale Public Domain MusicXML Dataset for Symbolic Music Processing}, 
  year={2025},
  volume={},
  number={},
  pages={1-5},
  doi={10.1109/ICASSP49660.2025.10890217}}

@InProceedings{olimpic,
author="Mayer, Ji{\v{r}}{\'i}
and Straka, Milan
and Haji{\v{c}}, Jan
and Pecina, Pavel",
editor="Barney Smith, Elisa H.
and Liwicki, Marcus
and Peng, Liangrui",
title="Practical End-to-End Optical Music Recognition for Pianoform Music",
booktitle="Document Analysis and Recognition - ICDAR 2024",
year="2024",
publisher="Springer Nature Switzerland",
address="Cham",
pages="55--73",
isbn="978-3-031-70552-6"
}

@InProceedings{smt,
author="R{\'i}os-Vila, Antonio
and Calvo-Zaragoza, Jorge
and Paquet, Thierry",
editor="Barney Smith, Elisa H.
and Liwicki, Marcus
and Peng, Liangrui",
title="Sheet Music Transformer: End-To-End Optical Music Recognition Beyond Monophonic Transcription",
booktitle="Document Analysis and Recognition - ICDAR 2024",
year="2024",
publisher="Springer Nature Switzerland",
address="Cham",
pages="20--37",
isbn="978-3-031-70552-6"
}

@misc{audiveris,
  author = {Bitteur, Hervé and {Audiveris Contributors}},
  title = {Audiveris},
  url = {https://github.com/Audiveris/audiveris},
  version = {5.4.0}, 
  year = {2025},
}

@Article{pipeline_omr_2001,
author={Bainbridge, David
and Bell, Tim},
title={The Challenge of Optical Music Recognition},
journal={Computers and the Humanities},
year={2001},
month={May},
day={01},
volume={35},
number={2},
pages={95-121},
issn={1572-8412},
doi={10.1023/A:1002485918032},
url={https://doi.org/10.1023/A:1002485918032}
}

@Article{pipeline_omr_2012,
author={Rebelo, Ana
and Fujinaga, Ichiro
and Paszkiewicz, Filipe
and Marcal, Andre R. S.
and Guedes, Carlos
and Cardoso, Jaime S.},
title={Optical music recognition: state-of-the-art and open issues},
journal={International Journal of Multimedia Information Retrieval},
year={2012},
month={Oct},
day={01},
volume={1},
number={3},
pages={173-190},
issn={2192-662X},
doi={10.1007/s13735-012-0004-6},
url={https://doi.org/10.1007/s13735-012-0004-6}
}

@article{omr_review,
author = {Calvo-Zaragoza, Jorge and Jr., Jan Haji\v{c} and Pacha, Alexander},
title = {Understanding Optical Music Recognition},
year = {2020},
issue_date = {July 2021},
publisher = {Association for Computing Machinery},
address = {New York, NY, USA},
volume = {53},
number = {4},
issn = {0360-0300},
url = {https://doi.org/10.1145/3397499},
doi = {10.1145/3397499},
journal = {ACM Comput. Surv.},
month = jul,
articleno = {77},
numpages = {35}
}

@misc{imslp,
  author       = {{Project Petrucci LLC}},
  title        = {{IMSLP} Petrucci Music Library},
  howpublished = {\url{https://imslp.org}},
  year = {2026}
}

@INPROCEEDINGS{OpenScoreStringQuartets,
    author={Gotham, Mark and Redbond, Maureen and Bower, Bruno and Jonas, Peter},
    title={The ``{OpenScore String Quartet}'' Corpus},
    year=2023,
    publisher={Association for Computing Machinery},
    address={New York, NY, USA},
    doi={10.1145/3625135.3625155},
    booktitle={Proceedings of the 10th International Conference on Digital Libraries for Musicology},
    pages={49--57},
}

@INPROCEEDINGS{OpenScoreLieder,
    author={Gotham, Mark Robert Haigh and Jonas, Peter},
    title={The {OpenScore Lieder} Corpus},
    pages={131--136},
    publisher={Humanities Commons},
    booktitle={Music Encoding Conference Proceedings 2021},
    year=2022,
    doi={10.17613/1my2-dm23},
}

@misc{paddleocr,
      title={{PaddleOCR} 3.0 Technical Report}, 
      author={Cheng Cui and Ting Sun and Manhui Lin and Tingquan Gao and Yubo Zhang and Jiaxuan Liu and Xueqing Wang and Zelun Zhang and Changda Zhou and Hongen Liu and Yue Zhang and Wenyu Lv and Kui Huang and Yichao Zhang and Jing Zhang and Jun Zhang and Yi Liu and Dianhai Yu and Yanjun Ma},
      year={2025},
      eprint={2507.05595},
      archivePrefix={arXiv},
      primaryClass={cs.CV},
      url={https://arxiv.org/abs/2507.05595}, 
}

@inproceedings{mupt,
 author = {Qu, Xingwei and bai, yuelin and MA, Yinghao and Zhou, Ziya and Lo, Ka Man and LIU, JIAHENG and Yuan, Ruibin and Min, Lejun and Liu, Xueling and Zhang, Tianyu and Du, Xeron and Guo, Shuyue and Liang, Yiming and Li, Yizhi and Wu, Shangda and Zhou, Junting and Zheng, Tianyu and Ma, Ziyang and Han, Fengze and Xue, Wei and Xia, Gus and Benetos, Emmanouil and Yue, Xiang and Lin, Chenghua and Tan, Xu and Huang, Wenhao and Fu, Jie and Zhang, Ge},
 booktitle = {International Conference on Learning Representations},
 editor = {Y. Yue and A. Garg and N. Peng and F. Sha and R. Yu},
 pages = {46753--46779},
 title = {{MuPT}: A Generative Symbolic Music Pretrained Transformer},
 url = {https://proceedings.iclr.cc/paper_files/paper/2025/file/73f6f8897896f7bda86ea7d1ebc1dc4f-Paper-Conference.pdf},
 volume = {2025},
 year = {2025}
}

%%%%%%%%%%%%%%%%%%%%%%%%%%%%%%%%%%%%%%%%%%%%%%%%%%%%%%%%%%%%

\appendix

\section{Pipeline Details}

\subsection{Context Management Mechanism}
\label{ap:context_management}

Our VLM is trained to model the probability $P(a_i \mid a_{<i},u_i)$, where $a_i$ represents the system-level ABC notation and $u_i$ denotes the segmented image of the $i$th musical system. Because the preceding context $a_{<i}$ can be arbitrarily long while Legato~2 is constrained to a maximum context window of 2,048 tokens, we require a mechanism to manage the sequence length. Specifically, the tokens fed into the decoder are arranged in the following format:
\begin{equation*}
    \underbrace{\texttt{<|bos|>} \ a_{<i}}_{\text{left-truncated}} \
    \texttt{<|bos|>} \ \texttt{<|image|>} \ \texttt{<|bos|>} \underbrace{a_i}_{\text{right-truncated}}
\end{equation*}

The \texttt{<|image|>} token is a special control token that implicitly instructs the cross-attention module to attend to the visual features, while the image input $u_i$ is fed directly into the encoder. We apply left-truncation to the preceding system-level ABC sequence, as the most recent context is the most relevant for predicting the current system. Specifically, $a_{<i}$ is left-truncated to half of the model's maximum context length (i.e., 1,024 tokens in Legato~2).
While this truncation could potentially degrade the model's performance, the impact is minimal because the most critical musical context is highly localized. Furthermore, the multi-page recognition results presented in \S\ref{subsec:multi-page} demonstrate robust performance on lengthy documents.

\subsection{Training Details}
\label{ap:training_detail}

\paragraph{YOLO} We fine-tuned an YOLOv8 medium model ($\sim$26 million parameters). Training used an input resolution of $800$ pixels, batch size $16$, and $100$ epochs. The optimizer used was AdamW, using a base learning rate of $0.002$, betas of $0.9$ and $0.999$, weight decay of $0.0005$, a linear learning-rate decay schedule with final LR factor $0.01$, and a 3-epoch warm-up. Mixed-precision training is enabled. Validation was performed after each epoch, and the best checkpoint was selected using validation performance. This model was trained on one Nvidia A100 GPU for 6 hours.

\paragraph{VLM} Following \citet{legato1}, our VLM adopts the \lm{Llama-3.2-11B-Vision}~\citep{mllama} architecture. We retain the pre-trained vision encoder but reduce the size of the decoder, training it from scratch. Since only the small decoder and the multimodal projection layer are trained, the total number of trainable parameters is $113.7$M. We employ a learning rate of $3\times 10^{-4}$ and a batch size of $32$. 
The model is trained on the PDMX-Synth training split for five epochs. 
Optimization is performed using the AdamW optimizer ($\beta_1 =0.9, \beta_2 = 0.99, \epsilon=10^{-6}$) paired with a cosine decay learning rate scheduler and a warm-up ratio of $0.03$. This model was trained with 180 Nvidia L40 GPU hours.

\subsection{Rule-Based ABC Converter}
\label{ap:abc_converter}

The ABC converter is a rule-based algorithm designed to translate between standard ABC notation and our system-level ABC representation.

During the processing of the PDMX-Synth training data, we convert standard ABC notation into a system-level format. 
Given the standard ABC notation $a^*$ , we first parse the composition-level header information and then extract individual voices from the content. 
For each voice, we segment the notation into systems according to the linebreak symbol (\texttt{\$}). 
Finally, for a given system, we merge its corresponding voices and append the layout information (\texttt{\%\%score}). 
The composition-level header is retained exclusively at the beginning of the first system, $a_1$.

During inference, after our VLM generates the system-level ABC representations, we convert them back into standard ABC notation. 
Specifically, we first parse the sequence of system-level outputs $\langle a_1, a_2, \cdots, a_s\rangle$.
We then retrieve the composition-level header from $a_1$. Subsequently, we extract the final layout information from the last system, $a_s$, and insert it into the composition-level header. Finally, for each voice, we concatenate the content across all systems.
Note that this merging process occasionally fails due to ill-formed syntax generated by the VLM. 
In such cases, we iteratively discard the terminal system-level ABC outputs until the merge succeeds.

\section{Audiveris Preprocessing}
\label{section:audiveris_preprocess}
For Audiveris-based optical music recognition, the original score images were preprocessed by resampling them to a higher effective resolution before transcription. The original sample images were tagged at approximately 72 DPI, which cannot be processed by Audiveris. 
Each image was therefore scaled toward a target effective resolution of 300 DPI. Note that since the input images were rasters, upscaling does not recover details that were not originally present. Rather, more pixels are added to increase DPI so that Audiveris can process it. To avoid exceeding Audiveris's practical page-size limits, output images were capped at 20 million pixels, which caused some large pages to be clamped to a slightly lower  effective DPI (250--300). 
This preprocessing was limited to resolution normalization; no deskewing, denoising, thresholding, or binarization was applied. The purpose was to improve Audiveris's ability to detect musical structure from already-rasterized inputs.

\section{Evaluating YOLO Segmentation and Robustness}
\label{sec:yolo_eval}
\subsection{YOLO Segmentation Performance}
\label{subsec:yolo_segmentation}
We evaluate our YOLO model on a test set of 204 images, sampled from the same data sources as the training set. The model achieved a validation bounding box mAP of 
\[
\text{mAP}_{50:95}=0.804
\]
with 
\[
\text{mAP}_{50}=0.990,\quad
\text{Precision}=0.990,\quad
\text{Recall}=0.994.
\]
Note that these metrics likely underestimate the true capability of the YOLO model, since we stretch the left and right side of each bounding box to segment the page, and so the YOLO model only needs to predict the top and bottom borders of the box. 

\subsection{Robustness to YOLO Errors}

\label{subsec:yolo_robustness}
To measure the robustness of our system against YOLO segmentation errors, we conduct the following experiments. For each page in the Rendered OpenScore String Quartets dataset, we artificially pick, with uniform probability, to either (1) merge two neighboring boxes, or (2) delete one of the identified bounding boxes. Then, we make our VLM predict the corrupted samples. We find that after the corruptions, Legato~2 was able to achieve an OMR-NED of  $47.4$, scoring better than Audiveris and Gemini 3.1 Pro, showing that even in the rare case of YOLO failure, Legato~2 can still transcribe with competitive performance. 

\section{System Prompts}
\label{section:system_prompts}
\subsection{Sheet Music Recognition}
We use the following prompts for evaluating Gemini 3 and GPT 5 on standard OMR benchmarks: 
\begin{Verbatim}[breaklines, breakanywhere]
You will be given an image of a sheet of music. 
Transcribe it into valid ABC 2.1 notation. Try your best to transcribe and make a reasonable guess if the image is not clear.
You will be given three in context examples of image-transcription pairs
Output only the ABC (no explanations), preferably inside a single ```abc fenced block.
IMPORTANT: NEVER give outputs like: 'Unable to transcribe from the provided image due to insufficient resolution/clarity.' If you can't tell, give your best guess. 
!!!ALWAYS OUTPUT VALID ABC, DO NOT GIVE ANY ENGLISH OUTPUT. GIVE YOUR BEST GUESS IF YOU ARE NOT SURE!!!

Transcribe this score to ABC.
{icl_1_img}
{ICL_1_ABC}


Transcribe this score to ABC.
{icl_2_img}
{ICL_2_ABC}


Transcribe this score to ABC.
{icl_3_img}
{ICL_3_ABC}


Transcribe this score to ABC.
{image}
\end{Verbatim}
\subsection{Sheet Music Understanding}
\subsubsection{MusiXQA} 
\label{section:system_prompts:musicxqa}
\paragraph{Baseline} %We mostly follow the original paper's prompting format for our investigation. 
As a starting point for our investigation, our baseline prompt follows the format introduced by \citet{musixqa}. 
%We make a few key changes, including \vcomment{I don't know what you do that makes this "mostly follow" and not entirely follow, so someone who does add that in here!} \brian{addressed}
\begin{Verbatim}[breaklines,breakanywhere]
You are an AI assistant specializing in Optical Music Recognition
(OMR) and Optical Character Recognition (OCR) for music sheets.
Your task is to accurately analyze images of music notation and
provide structured responses to visual question-answering (VQA)
tasks.
You will process printed music sheet images and answer both
OCR and OMR-related questions with high accuracy.
1 OCR-Based Tasks (Text Extraction)
- Extract the title and composer from the music sheet.
- Identify and extract the tempo marking (in BPM).
- Recognize and return the time signature.
- Extract explicitly labeled chord names from the sheet.
2 OMR-Based Tasks (Music Symbol Recognition)
- Identify the number and type of clefs (e.g., treble, bass).
- Count the number of bars (measures) in the music sheet.
- Recognize repeat sections based on notation symbols.
- Extract note durations (e.g., quarter, eighth, dotted notes, tied notes) for a given bar.
- Identify note pitches within a given bar.
- Return a structured representation of pitch, duration for a given bar in JSON string of list of python dictionaries without indent.
- Use kern representation for duration.
- If no explicit chord labels exist, infer the chord based on the notes in a given bar.
3 Response Format
- Provide structured, precise, and as concise as possible answers.
- Use structured JSON output without indent, when applicable for easy parsing.
4 Additional Considerations
- Ensure responses are notation-aware, considering key signatures, accidentals, and note relationships.
- Handle staff line separation correctly, ensuring multi-clef scores are properly analyzed.
- Avoid hallucinating missing information; only extract what is present in the image.
Follow music engraving conventions and OMR best practices to provide accurate, structured answers. If the requested information is not visible in the image, respond with '"Information not found"' instead of making assumptions.

Question:
{question}
"""
\end{Verbatim}
%\vcomment{does this prompt have the Question:
%{question} like the tool one does?}\brian{addressed}

\paragraph{Transcription as Context Provider} 
%Here, we provide the prompt given to VLMs when including the transcription of Legato/our model as optional context. \cite{legato1}
To investigate the capability of VLMs when given transcriptions from OMR models as additional context, we provide a transcription from either Legato~1 \citep{legato1} or Legato~2 as optional context. %removed reference to tool use to match the body of the paper
\begin{Verbatim}[breaklines,breakanywhere]
You are an AI assistant specializing in Optical Music Recognition
(OMR) and Optical Character Recognition (OCR) for music sheets.
Your task is to accurately analyze images of music notation and
provide structured responses to visual question-answering (VQA)
tasks.
You will process printed music sheet images and answer both
OCR and OMR-related questions with high accuracy.
1 OCR-Based Tasks (Text Extraction)
- Extract the title and composer from the music sheet.
- Identify and extract the tempo marking (in BPM).
- Recognize and return the time signature.
- Extract explicitly labeled chord names from the sheet.
2 OMR-Based Tasks (Music Symbol Recognition)
- Identify the number and type of clefs (e.g., treble, bass).
- Count the number of bars (measures) in the music sheet.
- Recognize repeat sections based on notation symbols.
- Extract note durations (e.g., quarter, eighth, dotted notes, tied notes) for a given bar.
- Identify note pitches within a given bar.
- Return a structured representation of pitch, duration for a given bar in JSON string of list of python dictionaries without indent.
- Use kern representation for duration.
- If no explicit chord labels exist, infer the chord based on the notes in a given bar.
3 Response Format
- Provide structured, precise, and as concise as possible answers.
- Use structured JSON output without indent, when applicable for easy parsing.
4 Additional Considerations
- Ensure responses are notation-aware, considering key signatures, accidentals, and note relationships.
- Handle staff line separation correctly, ensuring multi-clef scores are properly analyzed.
- Avoid hallucinating missing information; only extract what is present in the image.
Follow music engraving conventions and OMR best practices to provide accurate, structured answers. If the requested information is not visible in the image, respond with '"Information not found"' instead of making assumptions.

Question:
{question}
Optional generated transcription for this image:
{transcription}
You may use the transcription to answer the question. 
\end{Verbatim}

\paragraph{Judge Prompt} G-Acc is calculated using LLM-as-a-Judge techniques; we provide the prompt for the judge model here:
%Additionally, we provide the prompt for the judge model when calculating G-Acc.
\begin{Verbatim}[breaklines,breakanywhere]
You are grading a model answer for a music-sheet QA task.
Given Question, GroundTruth, and Prediction, output:
- 1 if Prediction is semantically correct for the Question
- 0 otherwise

Rules:
- Be strict on musically meaningful symbols (pitch letter, accidental, octave, duration, tie, chord root/quality).
- Ignore insignificant formatting differences (extra whitespace, minor punctuation/casing).
- Output only one character: 1 or 0.
\end{Verbatim}
\subsubsection{SSMR-Bench}
\paragraph{Baseline} We provide here the prompt used to test VLMs on SSMR-Bench. %\vcomment{not sure I understand this. Where are the in context examples below? And where are you omitting abc and full image? if it looks like the next prompt, make that explicit} \brian{addressed}
\begin{Verbatim}[breaklines, breakanywhere]
Answer this multiple-choice music reading question.
Output exactly one capital letter: A, B, C, or D.
Do not provide any explanation.
Question: {example['question']}
\end{Verbatim}
\paragraph{Transcription as Context Provider} As with MusiXQA (\S\ref{section:system_prompts:musicxqa}) we augment our baseline prompt with optional context in the form of transcriptions from Legato~1 \cite{legato1} or Legato~2.
%Here, we provide the prompt given to VLMs when including the transcription of Legato/our model as optional context. \cite{legato1}
\begin{Verbatim}[breaklines, breakanywhere]
Answer this multiple-choice music reading question.
Output exactly one capital letter: A, B, C, or D.
Do not provide any explanation.
You may optionally use the provided Legato transcriptions alongside the image(s).
The transcriptions may be imperfect, so prefer the image if they conflict.
Question: {example['question']}
ABC context:
{example['abc_context']}

Main score image:
[inline image: sheet_music_img]

Optional Legato transcription for main score:
{main_score_legato_transcription}

Choices:
A. {choice_A}
B. {choice_B}
C. {choice_C}
D. {choice_D}
\end{Verbatim}
In the case where the answer choices themselves contain images, we use the following prompt:
\begin{Verbatim}[breaklines, breakanywhere]
Answer this multiple-choice music reading question.
Output exactly one capital letter: A, B, C, or D.
Do not provide any explanation.
You may optionally use the provided Legato transcriptions alongside the image(s).
The transcriptions may be imperfect, so prefer the image if they conflict.
Question: {example['question']}
ABC context:
{example['abc_context']}

Main score image:
[inline image: sheet_music_img]

Optional Legato transcription for main score:
{main_score_legato_transcription}

Choices:

A.
[inline image: choice_A]
Optional Legato transcription for A:
{choice_A_legato_transcription}

B.
[inline image: choice_B]
Optional Legato transcription for B:
{choice_B_legato_transcription}

C.
[inline image: choice_C]
Optional Legato transcription for C:
{choice_C_legato_transcription}

D.
[inline image: choice_D]
Optional Legato transcription for D:
{choice_D_legato_transcription}
\end{Verbatim}

\section{Evaluation Metric}
\subsection{OMR Normalized Edit Distance~(OMR-NED)}
\label{subsec:omr_ned_eval_metric}

We employ the standard OMR-NED metric for evaluating OMR qualilty.
Compared to traditional metrics such as Symbol Error Rate, OMR-NED offers several benefits. Rather than comparing predicted and reference encodings purely as strings, OMR-NED evaluates discrepancies at the level of rendered musical notation, enabling assessment across a broad range of musical symbols and layout-dependent features. Specifically, the metric represents a composition as a sequence of measures $(m_1, m_2, \cdots, m_n)$ and computes the sequence edit distance to the ground truth. Each measure is defined as a set of musical symbols, and the distance between two measures is computed using a set edit distance. By quantifying the error as the normalized number of insertions and deletions required to transform a predicted sheet music into the reference, OMR-NED offers a robust and efficient measure of transcription quality.

\subsection{Embedded Text Evaluation Metric}
\label{subsec:text_eval_metric}

We evaluate embedded text recognition using different metrics depending on the output format of each model. For OMR systems that produce symbolic music, we first convert all predictions to ABC notation. We then extract textual fields from the ABC header, including title, composer, and inline textual annotations within the music body, which are represented in ABC as quoted strings. Location indicators are removed, and the resulting text items are concatenated in a canonical order using double underscores as separators. We compute character error rate (CER) between the predicted and reference strings using Levenshtein distance.

For OCR systems, outputs consist of unordered bounding boxes with associated text. Because these predictions cannot be reliably converted into the same canonical ordering, we evaluate them with an unordered character error rate (Unordered-CER). Specifically, we compute a set edit distance between predicted and reference text boxes, where inserting or deleting a box costs the length of its text and substituting one box for another costs their Levenshtein distance. This allows OCR outputs to be evaluated without imposing an arbitrary reading order.

\section{Additional Ablation Studies}

\subsection{Model Architecture}
\label{ap:model_arcitecture}
Following the Legato~1 architecture, our default model is initialized from \lm{Llama-3.2-11B-Vision} and configured with a reduced decoder size. We investigate the potential of scaling up the model capacity and exploring alternative architectures. 
Specifically, we examine: (i) whether a deeper or wider decoder enhances performance, and (ii) whether architectures utilizing different pretrained vision encoders yield improvements. To address (i), we construct two variants---one with double the number of layers, and another with a larger hidden dimension. For both variants, the total number of trainable parameters is controlled to approximately twice that of the default model. To address (ii), we adapt \lm{Qwen-3-VL-2B-Instruct}~\citep{bai2025qwen3vltechnicalreport} by reducing its decoder to match our default size, while maintaining an identical training pipeline.
As shown in Table~\ref{tab:architecture_ablations}, despite the much higher computational cost of the wide and deep decoder variants, only the deep decoder yields an improvement in performance over the default model. However, we note that this improvement is primarily localized to the PDMX-Synth test set, whereas gains on the remaining datasets are marginal. This discrepancy points to a distribution shift between the synthetic PDMX-Synth data and the real-world datasets, indicating that scaling up the decoder capacity largely results in better fitting the synthetic training distribution. Furthermore, the model with \lm{Qwen-3-VL-2B-Instruct} architecture underperforms the default model. We think that this is because its architecture assigns an individual image token to each small image patch, which rapidly exhausts the available context window, particularly when processing large images.

%We test whether architectural changes improve downstream performance by modifying both the decoder and the vision-language backbone. For the decoder, we evaluate deeper and wider variants with roughly twice the parameters of the original decoder. For the backbone, we adapt \lm{Qwen-3-VL-2B-Instruct} by reducing its decoder to match our 125M decoder budget while keeping the training pipeline unchanged. None of these variants yields meaningful gains, suggesting that architecture alone is not the main bottleneck; full evaluation details are provided in Appendix~\ref{}. \brian{add this later, @Guang can you check this section and potentially talk about what you did to make it wider?}

\begin{table}[tb]
\caption{\textbf{Ablation study on model architecture.} D-Dec and W-Dec refer to the Deep Decoder and Wide Decoder configurations. Qwen refers to using \lm{Qwen-3-VL-2B-Instruct} model architecture.
%\vcomment{results! I want the information in this caption while I'm reading the relevent section. At the very least you should point me to the table, but this is a very detailed caption for a table. My opinion is that a caption should not introduce new information, but should describe the figure and remind me of information in the paper maybe now contextualized differently}
}
\label{tab:architecture_ablations}
\centering
\begin{tabular}{lcccc}
\toprule
\multirow{2}{*}{Dataset (\# of Pages)}  & \multicolumn{4}{c}{OMR-NED $\downarrow$} \\
 & D-Dec & W-Dec & Qwen & Default\\
\midrule
{\em PDMX-Synth Test Set} (411)       &  $\mathbf{21.4}$ & $24.3$ & $25.6$ & $23.5$\\
{\em Rendered OpenScore String Quartets} (252) & $18.9$ & $22.6$ & $20.0$ & $\mathbf{17.1}$\\
{\em Camera OpenScore String Quartets} (252) & $33.2$ & $41.5$ & $43.7$ & $\mathbf{31.6}$ \\
{\em Rendered OpenScore Lieder} (64)        & $\mathbf{26.7}$ & $28.0$ & $29.4$ & $27.6$ \\
{\em Camera OpenScore Lieder} (64)          & $\mathbf{43.0}$ & $47.7$ & $45.0$ & $43.6$\\
{\em IMSLP Piano Scores} (32)           & $\mathbf{33.1}$ & $41.2$ & $38.4$ & $34.2$ \\
\bottomrule
\end{tabular}
\end{table}

\subsection{Data Distribution Shift}
\label{ap:data_distribution}
\begin{figure}[tb]
    \centering
    \includegraphics[width=\textwidth]{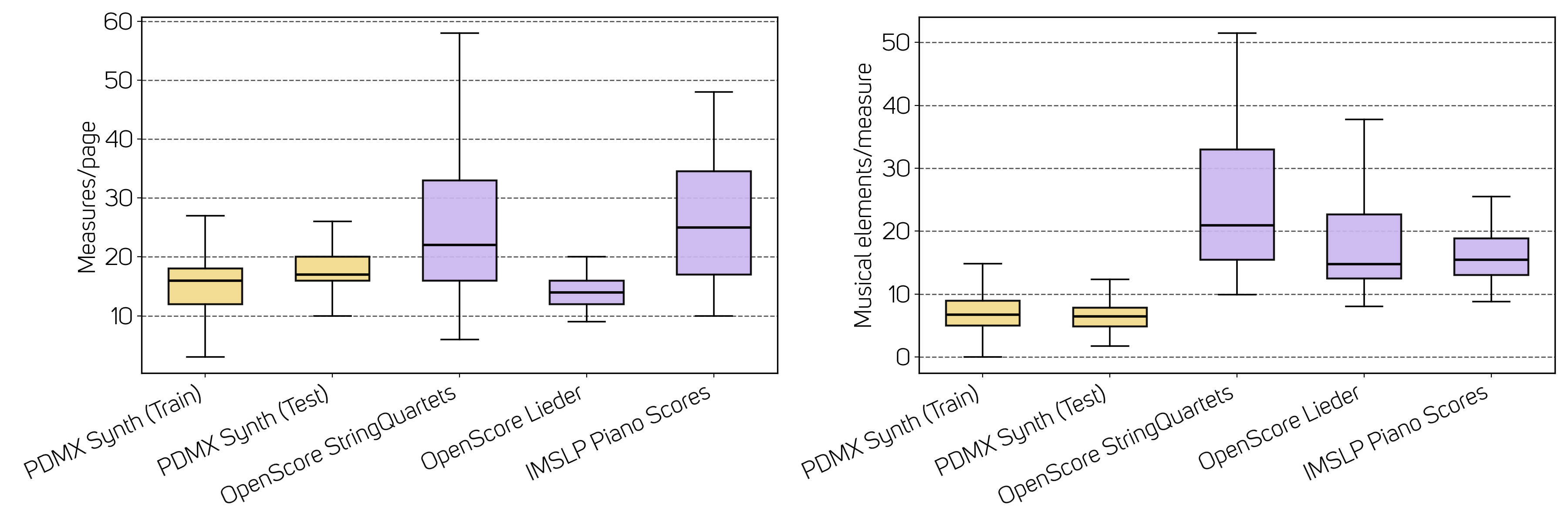}
    \caption{\textbf{Data distribution shift.} We plot the distribution of number of measures per page (left) and number of elements per measure (right).  \textcolor[RGB]{239,221,157}{\rule{1.5ex}{1.5ex}} denotes PDMX-Synth and \textcolor[RGB]{204,189,235}{\rule{1.5ex}{1.5ex}} denotes evaluation datasets. Sheet music in evaluation datasets is denser than PDMX-Synth. 
    %\textbf{(Left)} shows the distribution in measures per page for each rendered page in their respective datasets. \textbf{(Right)} shows the musical elements per measure. We find that although the both the train split and the test benchmarks shows similar measures per page, the three test benchmarks, OpenScore StringQuartets, OpenScore Lieder, and CameraIMSLP all show significantly higher musical elements per measure, making them harder for OMR systems to generalize to and potentially explaining why architecture changes struggle to improve on these tasks. 
    }
    \label{fig:distribution_shift}
    %\vcomment{this one is colored too, but it is also distracting to have the colored background (personal opinion). I do think that you should change the axes and labels to black for accessibility though, the purple text is harder on the eyes.}\brian{Addressed}
    
\end{figure}

To investigate the hypothesis that the distribution gap between the training and evaluation datasets hinders the performance gains of the scaled models, we plot the distributions of two key structural metrics across all datasets: (1) the number of measures per page, and (2) the number of musical elements per measure. We remove outliers, defined as any datapoint 1.5 times the inter-quartile range above or below the first or third quartile, before generating plots. 
As illustrated in Figure~\ref{fig:distribution_shift}, the evaluation datasets contain significantly more measures per page (with the exception of OpenScore Lieder, where vocal staves are masked, leaving blank spaces) and a higher density of musical elements per measure. Because PDMX-Synth is rendered from PDMX---a symbolic music dataset sourced from the MuseScore forum---the visual density of the resulting images is often low and sparse, frequently featuring empty pages or simplistic, single-voice compositions. 
This fundamental difference in content density accounts for the observed distribution shift. 
Consequently, we argue that the curation of large-scale, real-world datasets is the primary bottleneck for training more robust OMR models and narrowing this generalization gap.

\subsection{Naive BPE Tokenizer}
\label{ap:naive_tokenizer}

We present the evaluation results for a model utilizing a tokenizer naively trained on the text-inclusive PDMX-Synth dataset. The goal of this model was to achieve text awareness by directly applying Legato's training recipe. 
However, as shown in Table~\ref{tab:naive_tokenizer}, it performs significantly worse than both Legato~1 and 2, indicating that Legato's training approach is ineffective when directly applied to textual data.

\begin{table}[tb]
\caption{\textbf{Evaluation of model with naive BPE tokenizer.} Performances are significantly worse than both Legato~1 and our proposed models, indicating that Legato's training approach is ineffective when directly applied to textual data.}
\label{tab:naive_tokenizer}
\centering
\begin{tabular}{lc}
\toprule
Dataset & Raw 2048 Tokenizer \\
\midrule
{\em PDMX-Synth Test} & $48.3$ \\
{\em Camera OpenScore String Quartets} & $92.1$ \\
{\em Rendered OpenScore String Quartets} & $87.4$ \\
{\em Camera OpenScore Lieder} & $74.5$ \\
{\em Rendered OpenScore Lieder} & $64.4$ \\
{\em IMSLP Piano Scores} & $77.2$ \\
\bottomrule
\end{tabular}
\end{table}

\section{Validation Performances}
\label{ap:validation}

We detail the grid search conducted to identify the optimal inference hyperparameters for each of our ablation models, and our baseline. 
These hyperparameters are selected based on model performance on our manually curated validation set. 
Table~\ref{tab:val_legato_main} reports the validation OMR-NED for Legato~2. Table~\ref{tab:val_legato_main} reports the validation OMR-NED for our baseline Legato~1 model. Note that for Legato, the setting with the lowest validation loss also happened to be the setting reported in the original paper (10 beams with 1.1 repetition penalty)~\citep{legato1}. In the following subsections, we present the validation performance of all models evaluated in our ablation studies.

\begin{table}[htbp]
    \centering
    \caption{\textbf{OMR-NED of Legato~1 and Legato~2 on validation set.}}
    \label{tab:val_legato_main}
    \begin{subtable}[t]{0.45\textwidth}
        \centering
        \begin{tabular}{lccc} 
            \toprule
            & \multicolumn{3}{c}{Repetition Penalty} \\
            \cmidrule(lr){2-4} 
            \# Beams & $1.0$ & $1.1$ & $1.2$ \\
            \midrule
            $1$ &  $72.8$ & $71.2$ & $72.1$ \\
            $2$ & $70.7$ & $70.4$ & $70.4$ \\
            $5$ & $71.3$ & $70.9$ & $71.8$ \\
            $10$ & $71.8$ & $\textbf{70.2}$ &  $70.6$ \\
            \bottomrule
        \end{tabular}
        \caption{Legato}
    \end{subtable}
    \hfill
    \begin{subtable}[t]{0.45\textwidth}
        \centering
        \begin{tabular}{lccc} 
            \toprule
            & \multicolumn{3}{c}{Repetition Penalty} \\
            \cmidrule(lr){2-4} 
            \# Beams & $1.0$ & $1.1$ & $1.2$ \\
            \midrule
            $1$ &  $50.8$ & $\mathbf{46.8}$ & $56.2$ \\
            $2$ & $51.2$ & $49.0$ & $47.9$ \\
            $5$ & $54.5$ & $50.6$ & $51.6$ \\
            $10$ & $58.5$ & $54.3$ &  $52.6$ \\
            \bottomrule
        \end{tabular}
        \caption{Legato~2}
    \end{subtable}
\end{table}

\subsection{System Segmentation and Byte Fallback}

In our ablation studies concerning system segmentation and byte fallback, we report the performance of three models in the main text: Legato, Legato~1 with system segmentation (\texttt{SS}), and Legato~1 with both system segmentation and byte fallback (\texttt{SS} + \texttt{BF}). For the Legato~1 model, the validation performance is reported in Table~\ref{tab:val_legato_main}. For the other two models, we report the validation performance in Table~\ref{tab:val_ss}.

\begin{table}[htbp]
    \centering
    \caption{\textbf{OMR-NED of models for system segmentation and byte fallback on validation set.}}
    \label{tab:val_ss}
    \begin{subtable}[t]{0.45\textwidth}
        \centering
        \begin{tabular}{lccc} 
            \toprule
            & \multicolumn{3}{c}{Repetition Penalty} \\
            \cmidrule(lr){2-4} 
            \# Beams & $1.0$ & $1.1$ & $1.2$ \\
            \midrule
            $1$ &  $57.9$ & $49.6$ & $55.3$ \\
            $2$ & $50.7$ & $50.7$ & $49.4$ \\
            $5$ & $51.1$ & $50.8$ & $\mathbf{48.5}$ \\
            $10$ & $51.1$ & $51.2$ &  $48.8$ \\
            \bottomrule
        \end{tabular}
        \caption{Legato~1 + \texttt{SS}}
    \end{subtable}
    \hfill
    \begin{subtable}[t]{0.45\textwidth}
        \centering
        \begin{tabular}{lccc} 
            \toprule
            & \multicolumn{3}{c}{Repetition Penalty} \\
            \cmidrule(lr){2-4} 
            \# Beams & $1.0$ & $1.1$ & $1.2$ \\
            \midrule
            $1$ &  $57.1$ & $50.0$ & $52.6$ \\
            $2$ & $51.2$ & $49.9$ & $\mathbf{48.5}$ \\
            $5$ & $50.4$ & $50.2$ & $52.1$ \\
            $10$ & $55.1$ & $53.6$ &  $50.6$ \\
            \bottomrule
        \end{tabular}
        \caption{Legato~1 + \texttt{SS} + \texttt{BF}}
    \end{subtable}
    \hfill
\end{table}

\subsection{Vocabulary Size}

In Table~\ref{tab:val_vocab}, we report the validation OMR-NED scores for the model with a vocabulary size of 8,192, which corresponds to the rightmost column of Table~\ref{tab:ablation}. The model with a vocabulary size of 2,048 is identical to the ``Legato~1 + \texttt{SS} + \texttt{BF}'' configuration, and its validation errors are reported in Table~\ref{tab:val_ss}. Finally, the model with a vocabulary size of 4,096 represents our final model, whose validation errors are detailed in Table~\ref{tab:val_legato_main}. Note that all evaluated models utilize a text-aware tokenizer featuring byte fallback; the vocabulary size is the sole independent variable.

\begin{table}[htbp]
    \centering
    \caption{\textbf{OMR-NED of model with vocabulary size of 8,192 on validation set.}}
    \label{tab:val_vocab}
    \begin{tabular}{lccc} 
        \toprule
        & \multicolumn{3}{c}{Repetition Penalty} \\
        \cmidrule(lr){2-4} 
        \# Beams & $1.0$ & $1.1$ & $1.2$ \\
        \midrule
        $1$ &  $56.0$ & $54.4$ & $62.0$ \\
        $2$ & $55.5$ & $\mathbf{52.8}$ & $53.8$ \\
        $5$ & $56.2$ & $55.4$ & $55.3$ \\
        $10$ & $56.2$ & $54.9$ &  $54.5$ \\
        \bottomrule
    \end{tabular}
\end{table}

\subsection{Model Architecture}

Table~\ref{tab:val_architecture} presents the validation performance for the models used in the architecture ablation studies in Appendix~\ref{ap:model_arcitecture}.  The evaluated configurations include a deep decoder, a wide decoder, and the \lm{Qwen-3-VL-2B-Instruct} architecture. 
Consistent with our earlier approach, we select the hyperparameters that achieve the lowest validation OMR-NED.

\begin{table}[htbp]
    \centering
    \caption{\textbf{OMR-NED of models with various architecture on validation set.} D-Dec and W-Dec refer to the Deep Decoder and Wide Decoder configurations. Qwen refers to using \lm{Qwen-3-VL-2B-Instruct} model architecture.}
    \label{tab:val_architecture}
    \begin{subtable}[t]{0.45\textwidth}
        \centering
        \begin{tabular}{lccc} 
            \toprule
            & \multicolumn{3}{c}{Repetition Penalty} \\
            \cmidrule(lr){2-4} 
            \# Beams & $1.0$ & $1.1$ & $1.2$ \\
            \midrule
            $1$ &  $62.5$ & $59.9$ & $61.6$ \\
            $2$ & $61.1$ & $61.7$ & $61.1$ \\
            $5$ & $61.6$ & $\mathbf{59.8}$ & $60.1$ \\
            $10$ & $64.7$ & $62.8$ &  $60.7$ \\
            \bottomrule
        \end{tabular}
        \caption{Qwen}
    \end{subtable}
    \vfill
    \begin{subtable}[t]{0.45\textwidth}
        \centering
        \begin{tabular}{lccc} 
            \toprule
            & \multicolumn{3}{c}{Repetition Penalty} \\
            \cmidrule(lr){2-4} 
            \# Beams & $1.0$ & $1.1$ & $1.2$ \\
            \midrule
            $1$ &  $55.1$ & $52.1$ & $53.3$ \\
            $2$ & $51.6$ & $50.6$ & $51.8$ \\
            $5$ & $50.5$ & $51.4$ & $50.9$ \\
            $10$ & $55.1$ & $52.1$ &  $\mathbf{46.4}$ \\
            \bottomrule
        \end{tabular}
        \caption{D-Dec}
    \end{subtable}
    \hfill
    \begin{subtable}[t]{0.45\textwidth}
        \centering
        \begin{tabular}{lccc} 
            \toprule
            & \multicolumn{3}{c}{Repetition Penalty} \\
            \cmidrule(lr){2-4} 
            \# Beams & $1.0$ & $1.1$ & $1.2$ \\
            \midrule
            $1$ &  $60.0$ & $\mathbf{55.3}$ & $\mathbf{55.3}$ \\
            $2$ & $56.2$ & $55.9$ & $\mathbf{55.3}$ \\
            $5$ & $58.3$ & $\mathbf{55.3}$ & $\mathbf{55.3}$ \\
            $10$ & $61.8$ & $59.6$ &  $57.5$ \\
            \bottomrule
        \end{tabular}
        \caption{W-Dec}
    \end{subtable}
\end{table}

\section{Qualitative Examples}

\subsection{Sheet Music Understanding}
\label{subsec:sheet_music_understanding}
\begin{figure}[tb]
    \centering
    \includegraphics[width=\linewidth]{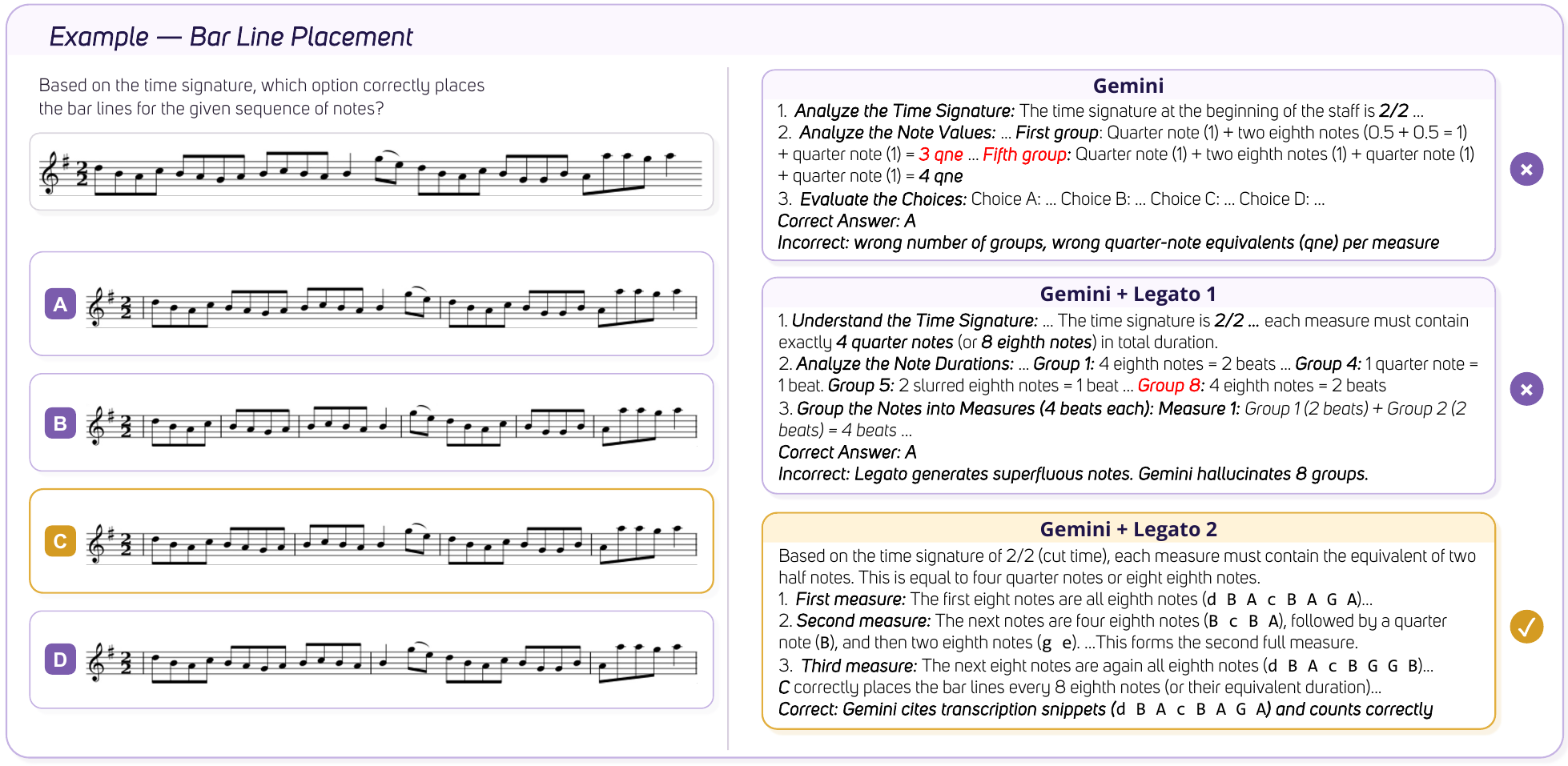}
    \caption{\textbf{Qualitative example from SSMR-Bench.} Left: The question and corresponding choices, both accompanied by sheet music. Right: Gemini's reasoning process and final answer under different context providers.}
    \label{fig:smu_example}
\end{figure}

In Figure~\ref{fig:smu_example}, we present an example from SSMR-Bench alongside Gemini's reasoning process under different context-provider settings. 
Note that for this example, we ask Gemini to output its reasoning process, instead of directly outputting the answer like we do during the actual evaluation on SSMR-Bench. 
The prompt asks for the correct barline placement within the provided sheet music. 
When relying exclusively on the image, Gemini successfully recognizes the time signature but fails to extract more fine-grained visual details. Consequently, it hallucinates five note groups, assigning an arbitrary number of beats to each. 
As for Legato~1, in this specific example, it generates superfluous notes at the end of its transcription. As a result, when Legato~1 serves as the context provider, Gemini is misled; it hallucinates eight groups, although it correctly calculates the durations for them (e.g., groups 4 and 5). 
In contrast, Legato~2 accurately transcribes the sheet music into ABC notation. By grounding its reasoning in this precise symbolic transcription, Gemini correctly parses the groups and computes the proper beat placements. 
In general, this example demonstrates how Legato~2 effectively provides reliable symbolic context to assist frontier language models in downstream sheet music understanding tasks.
%We provide an example evaluation of Gemini's response to a sample of SSMR-Bench in Figure \ref{fig:smu_example}. We show that without any transcription, the baseline model hallucinates the number of note groups. For this particular image, Legato was not able to produce a good transcription, and so Gemini tries to read the number of groups directly from the image and hallucinates again. With higher-quality transcriptions from our model, Gemini directly quotes the provided transcription and was able to answer this task correctly. 

\section{Limitations and Future Work}
In this paper, we introduced a novel OMR approach that segments sheet music into individual systems and processes them autoregressively. This method achieves state-of-the-art performance in both single- and multi-page recognition, while uniquely integrating the extraction of embedded textual metadata. Furthermore, we demonstrated that Legato~2 effectively serves as an upstream context provider, supplying precise symbolic data to general-purpose VLMs to enhance their reasoning on domain-specific musical tasks.

Despite these advances, Legato~2 has notable limitations. First, resource constraints precluded extensive ablation studies on individual pipeline components. Specifically, we have yet to formally quantify how segmentation errors from the YOLO-based detector propagate and affect downstream reasoning when the OMR output is used as VLM context. Second, model performance is constrained by distribution shifts between synthetic training data and the diverse reality of physical sheet music, which varies widely in engraving style, scan quality, historical notation, and handwritten markings. These domain shifts can degrade both transcription accuracy and the reliability of the resulting symbolic context.

To address these limitations, future work must broaden training distributions and establish diagnostic benchmarks to precisely measure error propagation in downstream tasks. Ultimately, our findings point toward a highly effective, modular architecture for sheet music understanding. Rather than relying on monolithic general-purpose models, we envision a framework where diverse, specialized models—including OMR, audio transcription, and structural analysis—act as integrated context providers, assisting frontier VLMs in solving complex, open-ended musical problems.

Additionally, from a broader societal perspective, better OMR systems may also increase the risk of music piracy by making it easier to convert copyrighted sheet music scans or photographs into clean, editable symbolic formats that can be redistributed, modified, or rendered into new editions. This could reduce the ability of composers, publishers, and archives to control access to copyrighted works, especially if OMR tools are integrated into large-scale digitization or search pipelines. To mitigate this risk, releases should include clear acceptable-use policies, copyright-aware dataset curation, provenance tracking for generated transcriptions, and mechanisms such as watermarking or metadata retention where appropriate. Platforms deploying OMR at scale could also limit bulk transcription of copyrighted materials, support takedown workflows, and encourage use on public-domain, licensed, or user-owned scores.

\end{document}